\documentclass[10pt,twocolumn,letterpaper]{article}

\usepackage[pagenumbers]{cvpr} % To force page numbers, e.g. for an arXiv version

\definecolor{cvprblue}{rgb}{0.21,0.49,0.74}
\usepackage[pagebackref,breaklinks,colorlinks,allcolors=cvprblue]{hyperref}
\usepackage{multirow}
\usepackage{multicol}
\usepackage{colortbl}
\usepackage{xcolor}
\usepackage{graphicx}  
\usepackage{subcaption} 
\usepackage{booktabs}
\usepackage{float}

\def\paperID{4617} % *** Enter the Paper ID here
\def\confName{CVPR}
\def\confYear{2025}

\title{M$^3$-VOS: Multi-Phase, Multi-Transition, and Multi-Scenery \\Video Object Segmentation}
\author{
Zixuan Chen\footnotemark[1] \\
{\tt\small m13953842591@sjtu.edu.cn}  
\and
Jiaxin Li\footnotemark[1] \\
{\tt\small li\_jiaxin@sjtu.edu.cn}
\and
Liming Tan \\
{\tt\small spinningfever@sjtu.edu.cn}
\and
Yejie Guo \\
{\tt\small gyj123@sjtu.edu.cn}
\and
Junxuan Liang \\
{\tt\small whitefork@sjtu.edu.cn}
\and 
Cewu Lu\\
{\tt\small lucewu@sjtu.edu.cn}
\and
Yong-Lu Li\footnotemark[2]\\
{\tt\small yonglu\_li@sjtu.edu.cn} \\
Shanghai Jiao Tong University \\
}

\begin{document}
\maketitle

\renewcommand{\thefootnote}{\fnsymbol{footnote}}
\footnotetext[1]{Equal contributions. $\dagger$Corresponding author.}
\renewcommand{\thefootnote}{\arabic{footnote}}

\begin{abstract}
Intelligent robots need to interact with diverse objects across various environments. 
The appearance and state of objects frequently undergo complex transformations depending on the object properties, \textit{e.g.}, phase transitions.
However, in the vision community, segmenting dynamic objects with phase transitions is overlooked.
In light of this, we introduce the concept of \textbf{phase} in segmentation, which categorizes real-world objects based on their visual characteristics and potential morphological and appearance changes. 
Then, we present a new benchmark, \textbf{Multi-Phase, Multi-Transition, and Multi-Scenery Video Object Segmentation (M$^3$-VOS)}, to verify the ability of models to understand object phases, which consists of \textbf{479} high-resolution videos spanning over 10 distinct everyday scenarios. It provides dense instance mask annotations that capture both object phases and their transitions.
We evaluate state-of-the-art methods on M$^3$-VOS, yielding several key insights. 
Notably, current appearance-based approaches show significant room for improvement when handling objects with phase transitions. The inherent changes in disorder suggest that the predictive performance of the forward entropy-increasing process can be improved through a reverse entropy-reducing process. These findings lead us to propose ReVOS, a new plug-and-play model that improves its performance by reversal refinement.
Our data and code will be publicly available at \url{https://zixuan-chen.github.io/M-cube-VOS.github.io/}.
\end{abstract}

\section{Introduction}
\label{sec:intro}

\begin{figure*}[t]
  \centering
   \includegraphics[width=0.97\linewidth]{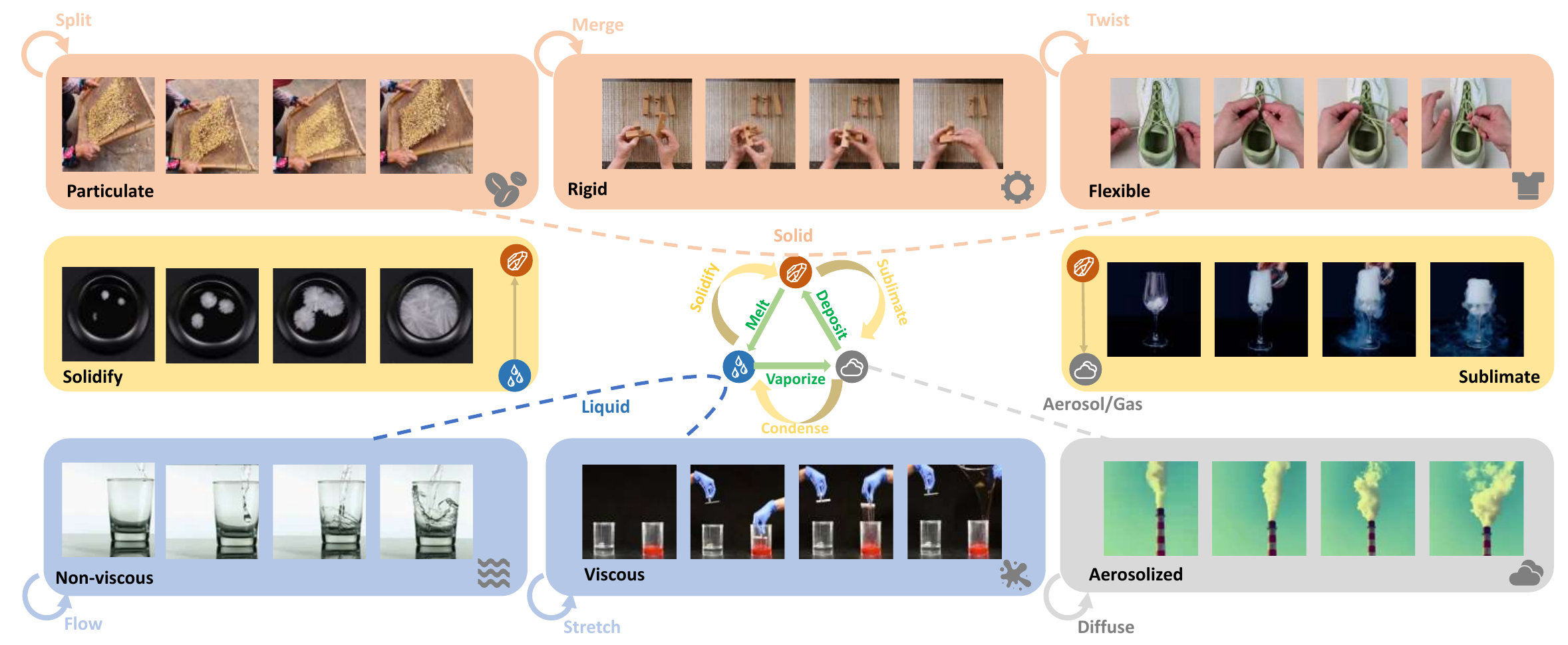}
   \vspace{-5px}
   \caption{Objects in the real world have various phase states. Each phase has its unique transition (intra-phase transition), such as water flowing, shoelace twisting, smoke diffusing, and so on. Besides, numerous transitions occur across different phases (cross-phase transition), such as bromine liquefaction, dry ice sublimation, and water freezing.}
   \label{fig:teaser}
   \vspace{-7px}
\end{figure*}

Object understanding is crucial, especially for Embodied AI. Lately, large-scale datasets and data-driven methods ~\cite{deng2009imagenet,lin2014microsoft} have advanced research in object understanding, shifting from traditional category recognition~\cite{krizhevsky2012imagenet} to deeper levels of comprehension, such as identifying higher-level  affordances~\cite{gibson1978ecological} and attributes~\cite{isola2015discovering}, which help establishing the concepts of objects in interactions with environment.

In the real world, intelligent robots need to interact with various objects in diverse ways, during which, objects exhibit a range of morphological and appearance changes influenced by their inherent characteristics, the nature of the environment, and the specific interactions occurring. 
Particularly, the \textbf{phase} characteristics of objects describe their intrinsic morphological features and change dynamics, making them an important component of object knowledge. 

Objects with different phase characteristics exhibit distinct change features when subjected to the same interactions. For instance, placing solid coffee beans in a blender results in their transformation into a powdered form. In contrast, pouring milk into the blender causes it to splash and flow continuously during blending, potentially even leading to separation at the end.
In addition to transformations \textbf{within} phase states, \textbf{cross}-state phase changes are common in the real world, endowing objects with properties they originally did not have. 
For example, dry ice rapidly sublimates, producing a lot of mist and water droplets will quickly frost at lower temperatures as \cref{fig:teaser}. Therefore, a deep understanding of an object's intrinsic phase characteristics and the properties of phase transition is crucial for robots to perceive the world, manipulate objects, and fulfill various complex functions according to human needs.

Despite efforts exploring the visual understanding of object change characteristics from various perspectives, such as OCL~\cite{li2023beyond} which focuses on changes in object usability, VOST~\cite{tokmakov2023breaking} and VSCOS~\cite{yu2023video} which emphasize the segmentation of objects with appearance variations, most works remain limited to a single phase. 
They often overlook the understanding of object knowledge across phase transitions. 

Thus, to address the absence of object phase transition understanding, we first categorize different objects based on their visual features and change characteristics commonly observed in daily life. 
Furthermore, we introduce \textbf{M$^3$-VOS}, a fine-grained text-visual annotated dataset for object knowledge understanding across multiple scenarios, including objects that encompass a variety of transformations, including appearance changes within phases and state transitions between phases.
Motivated by SA-V~\cite{ravi2024sam}, which was constructed using the pipeline combining SAM2-assisted manual annotation with SAM2-automatic generation, we designed our semi-annotate tool based on the Interactive Demo of Cutie~\cite{cheng2024putting}. Besides, we  propose an effective method of model cross-validation to mitigate model bias in model-assisted annotation tools.

Given the data, we analyze the current understanding of object state changes in the field of computer vision from the perspective of \textit{Video Object Segmentation (VOS}). 
On one hand, 
VOS task

requires the visual model to possess a comprehensive understanding of object phase knowledge to track objects that undergo significant changes in appearance and morphology during phase transitions. 
On the other hand, video object segmentation serves as an upstream task for other tasks, such as recognizing higher-level attributes and affordances and object manipulation tasks.

Equipped with our benchmark, we analyze state-of-the-art VOS algorithms~\cite{cheng2022xmem,cheng2024putting,zhou2024rmem,yang2022decoupling, ravi2024sam}. We found significant room for improvement in the understanding of objects undergoing phase transitions. 

Additionally, by analyzing the performance differences across cases, we identified potential challenges in phase changes and proposed specific improvements, establishing a baseline \textbf{ReVOS} utilizing reverse memory, booster, and readout fusion to advance object segmentation with phase transformations.

We conclude in \cref{sec:phase_performance} by analyzing the different performances of the existing models facing different categories of phase transitions. We hope this work will motivate further exploration into more robust object knowledge understanding. 

In summary, our main contributions are as follows:
\begin{enumerate}
\item We build a video object segmentation benchmark focused on object phase transitions, addressing the gap in understanding object phase transformations in this field.

\item We propose an efficient semi-automated annotation tool tailored for multi-phase object segmentation and a dual-model cross-validation method to address the common issue of model bias in model-assisted annotation tools.

\item To address the limitations of mainstream appearance-first models in understanding object phase transitions, we propose a bidirectional propagation module that enhances the model's performance in this task.
\end{enumerate}

\section{Related Work}
\label{sec:related_work}

%-------------------------------------------------------------------------

\subsection{Object Knowledge Understanding}
 
Recently, computer vision has seen significant advancements in object understanding, primarily represented by two paradigms. 
The first is the classification paradigm, which has evolved from object recognition~\cite{krizhevsky2012imagenet} to a deeper understanding of object attributes~\cite{isola2015discovering} and affordances~\cite{gibson1978ecological}. 
The second focuses on pixel-level classification, encompassing tasks such as image grounding~\cite{plummer2016flickr30kentitiescollectingregiontophrase,mao2016generationcomprehensionunambiguousobject} and segmentation of changing objects in videos~\cite{ravi2024sam,cheng2024putting}.
The former emphasizes an explicit understanding of object knowledge, where the model is expected to extract relevant information about objects directly from images~\cite{krizhevsky2012imagenet,gibson1978ecological,isola2015discovering}. 
In contrast, the latter highlights an implicit understanding of object knowledge, where the model determines the location of objects in an image based on given prompts and its implicit recognition of object knowledge~\cite{plummer2016flickr30kentitiescollectingregiontophrase,mao2016generationcomprehensionunambiguousobject,ravi2024sam,cheng2024putting}.

Data-driven foundation models~\cite{liu2024improvedbaselinesvisualinstruction} have achieved remarkable performance in both above paradigms and even demonstrate strong open-vocabulary abilities. However, very few studies focus on understanding object phase transitions, an essential aspect of object knowledge. There is a notable lack of effective benchmarks for testing this task.

In this work, we concentrate on object phases and the corresponding \textbf{intra-phase} and \textbf{cross-phase transitions}. We construct a benchmark for the second paradigm of object knowledge understanding, designed to evaluate the capabilities of current vision models in understanding object phase transitions. This benchmark aims to fill the gap and provide a platform to test models on this critical task.

\subsection{Video Object Segmentation}

\begin{table}[t]
    \centering
    \resizebox{\linewidth}{!}{
    \begin{tabular}{ccccccccc}
        \toprule
        Dataset                     & Videos &  Frames  & Avg len. (s) & Ann fps.  & Text & Fluid & CPT \\
        \midrule 
        DAVIS'16        \cite{perazzi2016benchmark}          & 20    &   1376  & 2.87               & 24     & $ \checkmark ^{\dag} $ & ×  & ×  \\
        DAVIS'17     \cite{Pont-Tuset_arXiv_2017}               & 90    & 6,265    & 2.90              & 24     & $ \checkmark ^{\dag} $ & ×  &  ×  \\
        YTVOS     \cite{xu2018youtubevossequencetosequencevideoobject}                  & 982  &  25,812 & 4.38              & 6      & $ \checkmark ^{\dag} $ & ×  &  ×  \\
        VSCOS     \cite{yu2023video}                   & 98   &   44036  &       8.34       & 0.35 $^*$   & ×                    & ×  & ×  \\
        VOST     \cite{tokmakov2023breaking}                     & 141 &  15,617    &    22.15            & 5      & ×                    & ×  &  × \\
        % MeVis                        & 294  &   &              & ?      & $ \checkmark $         & ×  & ×  \\
        \midrule
        M$^3$-VOS (Ours) & 479  &  205,181  & 14.27 &  30   & $ \checkmark $   &           $ \checkmark $ & $ \checkmark $ \\
        \bottomrule
    \end{tabular}}
    \vspace{-5px}
    \caption{Statistics of test set and val set in representative video segment datasets ($\dag$: DAVIS$_{16}$-RVOS, DAVIS$_{17}$-RVOS~\cite{khoreva2019video}, Refer-YouTube-VOS~\cite{seo2020urvos} was built based on these datasets; *: VSCOS only annotated some frames in the valid set; Fluid: liquid or aerosol/gas phase); \textbf{CPT: Cross-Phase Transitions}. M$^3$-VOS focuses on different phase transitions in multi-scenes,  motivating our design to have a higher video annotation frame rate.}
    \vspace{-5px}
    \label{tab:compare_benchmark}
\end{table}

Video object segmentation is the process of segmenting target foreground objects from the video background at the pixel level within a video sequence.
The target foreground object is typically specified either by providing a mask in the first frame, known as semi-supervised VOS~\cite{perazzi2016benchmark,xu2018youtubevossequencetosequencevideoobject,Pont-Tuset_arXiv_2017}.

This task requires the effective tracking of the target and a clear understanding of its boundaries, which demands that models have a thorough comprehension of the object’s appearance~\cite{perazzi2016benchmark,xu2018youtubevossequencetosequencevideoobject}, dynamic properties~\cite{ tokmakov2023breaking,yu2023video}, and motion information~\cite{ding2023mevis} within the video, as well as other physical characteristics. This is precisely why we chose video object segmentation as the specific task to evaluate the ability of vision models to understand object phase transitions.

To prompt the development of Video Object Segment,  \cite{perazzi2016benchmark,xu2018youtubevossequencetosequencevideoobject,Pont-Tuset_arXiv_2017} constructs a series of early VOS benchmarks. Recently, several works have explored testing the open-world capabilities of VOS models from different perspectives: VOST~\cite{tokmakov2023breaking} and VSCOS~\cite{yu2023video} focus on changes in object appearance; MeVis~\cite{ding2023mevis} focuses on the movement characteristics of objects. However, compared to the M$^3$-VOS dataset, these models are all limited to single-phase objects, typically dealing with solid objects. These objects have relatively simple transformation characteristics and highly predictable movement patterns.

There are a lot of VOS methods achieve success and can be concluded into two classes:

\textbf{Memory-based VOS}.
Semi-supervised Video Object Segmentation (SVOS) involves inference based on a ground-true mask for the first frame, and it computes the masks of the rest frames. The most frequently used methods are memory-based, which means they store pixel-level or object-level features in their memory bank and compare them with features from newly arrived frames. Recent methods~\cite{cheng2021rethinking, cheng2022xmem} improve the memory bank by stratifying the memory bank into short and frequently updated memories and long and infrequently updated memories to better balance efficiency and performance. Recent method Cutie \cite{cheng2024putting} introduces object memory, which is fused with pixel memory in an object transformer.

\textbf{Reverse propagation in VOS}. 
We proposed a method that adopts a refinement module and optimizes the segmentation by reversely propagating the mask through the video sequence. Among the previous methods, there are not many methods that use backward mask propagation. The most similar method is DyeNet~\cite{li2018video}, which uses a re-identification module to predict a set of high-munificence masks and uses a bi-directionally proper, nation to the rest frames. Other methods that include elements of reverse optimization typically occur in bidirectional settings for sequences, \eg bi-directional ConvLSTMs~\cite{song2018pyramid}. In contrast, Our method doesn't need a high-confidence mask and is implemented as a plug-and-play module as long as the backbone is a mask propagation-based architecture.

%-------------------------------------------------------------------------

\section{Preliminary}

In this section, we introduce the concept of phase and use it to classify everyday materials. Then, we categorize the transformations of objects into \textbf{Intra-Phase Transitions} and \textbf{Cross-Phase Transitions}.

\subsection{Phase Categories}

Rather than defining the phases of objects from the microscopic perspective provided by chemistry or physics, which focuses on molecular spacing, we adopt a \textit{macroscopic} approach to classify objects based on their appearance and dynamic characteristics. In our study, phase refers to an attribute that describes both visual characteristics and intrinsic transformation properties. 
Next, we categorize commonly encountered objects into three major types: \textit{solid, liquid}, and \textit{aerosol/gas}. Then we introduce the subdivided phase categories for each major class of materials, along with their respective transformation properties (Tab.~\ref{table:Phase}).

\begin{table}[t] 
\small 
\centering
\resizebox{0.95\linewidth}{!}{
\begin{tabular}{lcc|cr}
\toprule
\multicolumn{3}{l|}{Phase}                                                                                           & \multicolumn{2}{l}{Intra-Phase Transition}  \\ 
\bottomrule
\multicolumn{1}{l|}{\multirow{3}{*}{Solid}}  & \multicolumn{2}{l|}{Particulate}                                      & \multicolumn{2}{l}{Split}                   \\ \cline{2-5} 
\multicolumn{1}{l|}{}                        & \multicolumn{1}{l|}{\multirow{2}{*}{Non-Particulate}} & Rigid Body    & \multicolumn{2}{l}{Separate, Merge}        \\ \cline{3-5} 
\multicolumn{1}{l|}{}                        & \multicolumn{1}{l|}{}                                 & Flexible Body & \multicolumn{2}{l}{Twist, Stretch}          \\ \hline
\multicolumn{1}{l|}{\multirow{2}{*}{Liquid}} & \multicolumn{2}{l|}{Viscous}                                          & \multicolumn{2}{l}{Stretch, Paint}          \\ \cline{2-5} 
\multicolumn{1}{l|}{}                        & \multicolumn{2}{l|}{Non-Viscous}                                      & \multicolumn{2}{l}{Flow, Mix, Split, Paint} \\ \hline
\multicolumn{3}{l|}{Aerosol / Gas}                                                                                   & \multicolumn{2}{l}{Diffusion}               \\ 
\bottomrule
\end{tabular}}
\vspace{-5px}
\caption{Phase definition and intra-phase transition according to visual characteristics and intrinsic transformation properties. } 
\vspace{-5px}
\label{table:Phase}
\end{table}

For \textbf{solid materials}, particle size also influences their transformation characteristics. Therefore, we further subdivide solid materials into fragmentary or powder-like substances and non-fragmentary substances. For non-fragmentary substances, we categorize them further based on whether the material can easily undergo arbitrary deformation, dividing them into rigid and flexible objects.

For \textbf{liquid materials}, substances like melted cheese or melted rubber do not frequently exhibit fluidity in everyday situations. Instead, they demonstrate viscous properties, such as stringiness, when interacting with other materials. Therefore, we classify liquid materials based on their degree of fluidity and viscosity, dividing them into viscous and non-viscous substances.

For \textbf{aerosols/gases}, it not only includes common gaseous substances but also solid particles and liquid droplets suspended in the air. These materials possess the property of continuous diffusion in the atmosphere.

\subsection{Phase Transition}

In everyday life, objects undergo various changes through interactions with the environment or humans. Based on whether these changes cross different phase states, we categorize them into \textit{intra-phase} and \textit{cross-phase} transitions.

\textbf{Intra-Phase Transformations} primarily depend on the transformation characteristics inherent to the phase state of the object itself. Different phase states of materials often exhibit unique intra-phase transformations. 
For example, flexible solids can be twisted (\eg, shoelaces or knots), viscous substances can stretch (\eg, melted cheese or syrup), and aerosols/gases can diffuse (smoke or steam diffusion).

\textbf{Cross-Phase Transformations} are quite common in everyday life but are often overlooked. Due to the differences in phase states before and after the transformation, these objects exhibit distinctly different visual characteristics and transformation properties. Therefore, understanding such transformations poses a greater challenge.

In Supplementary, we provide more detailed definitions of these phases and phase transition.
\section{Dataset Design and Construction}
\label{sec:dayaset_design}

In this section, we introduce the data collection and annotation of M$^3$-VOS. The key steps include selecting representative videos, annotating them with instance masks and information text, and defining an evaluation protocol.

\subsection{Video Collection} 

\begin{figure*}[htbp]
    \centering
    \begin{minipage}{0.68\textwidth}
     \begin{subfigure}{\textwidth} 
        \centering
        \includegraphics[width=\linewidth]{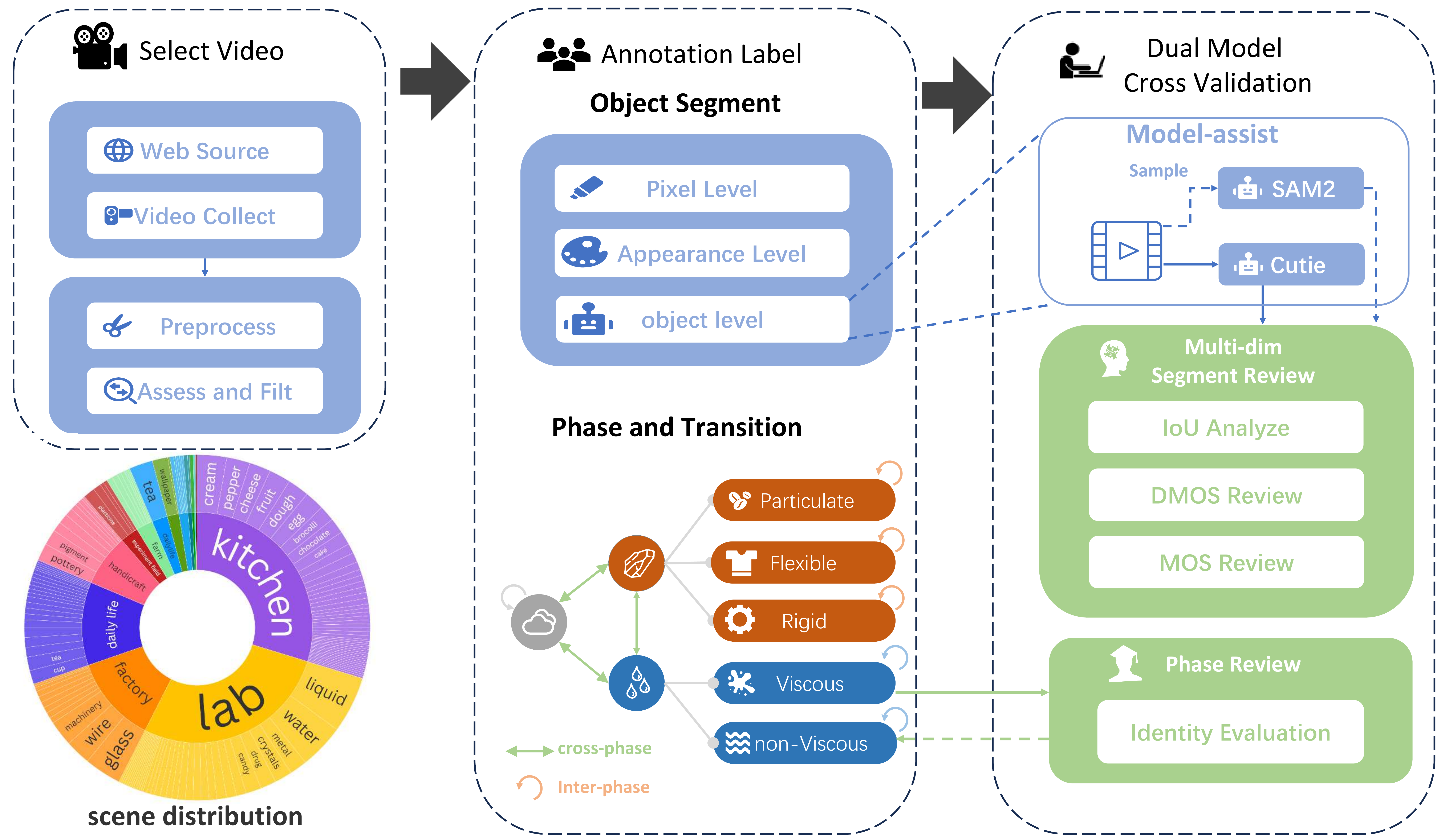} 
        \caption{Dataset construction.
        The core pipeline component is a multi-level semi-auto segmentation tool including Eraser Brush (\textbf{pixel-level}), 
        Color Difference Mask (\textbf{appearance-level}), 
        and Model-Assisted Annotation (\textbf{object-level}), 
        enabling the efficient annotation of objects with complex phase transitions.}
        \label{fig:dataset_pipeline}
       \end{subfigure}
    \end{minipage}%
    \hfill
    \begin{minipage}{0.32\textwidth}
        \begin{subfigure}{\textwidth} 
            \centering
            \includegraphics[width=\textwidth]{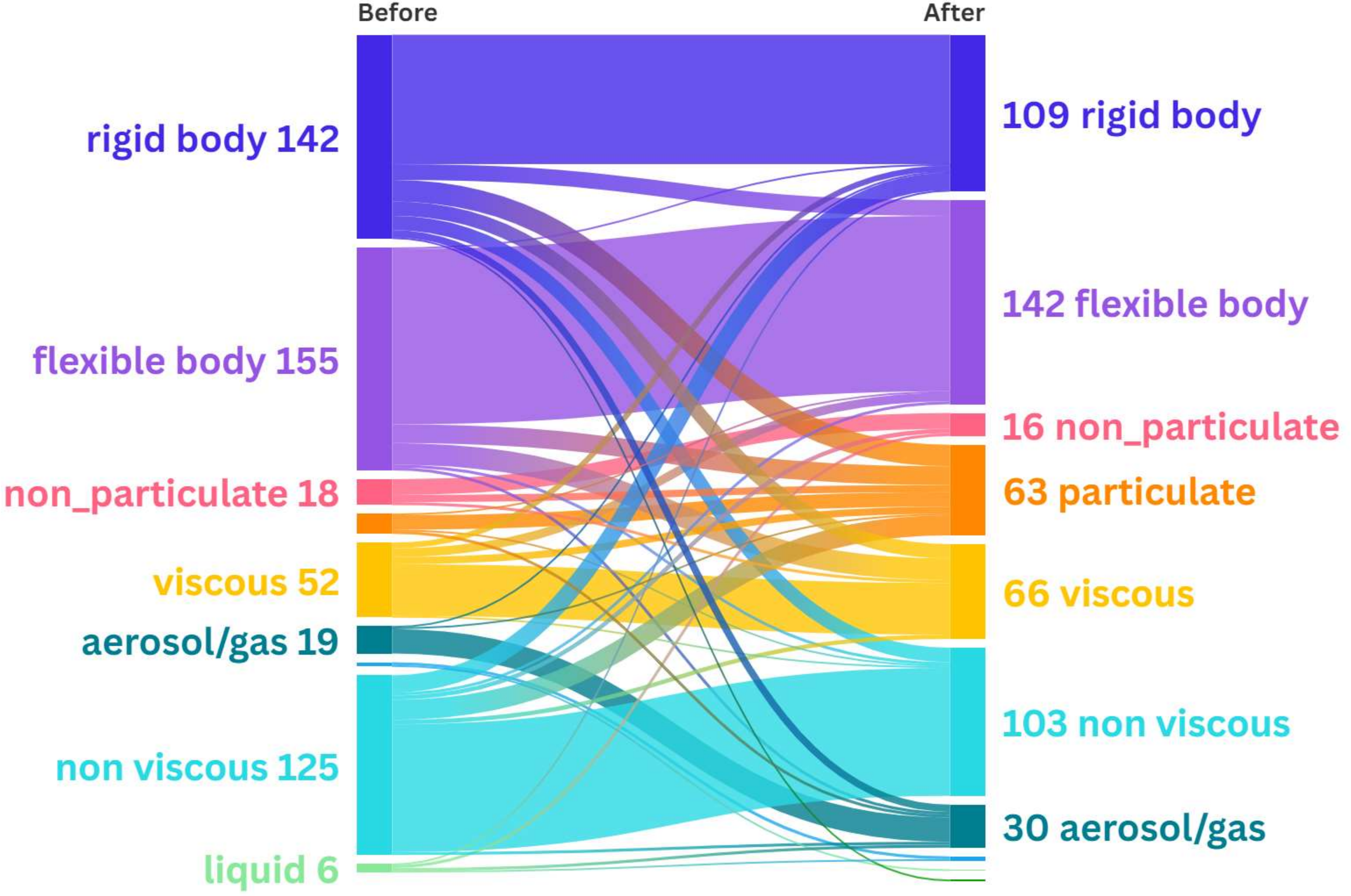} 
         \caption{Transition of phase.}
            \label{fig:sankey_before_after}
        \end{subfigure}\\[1ex] 
        \begin{subfigure}{\textwidth} 
            \centering
            \includegraphics[width=\textwidth]{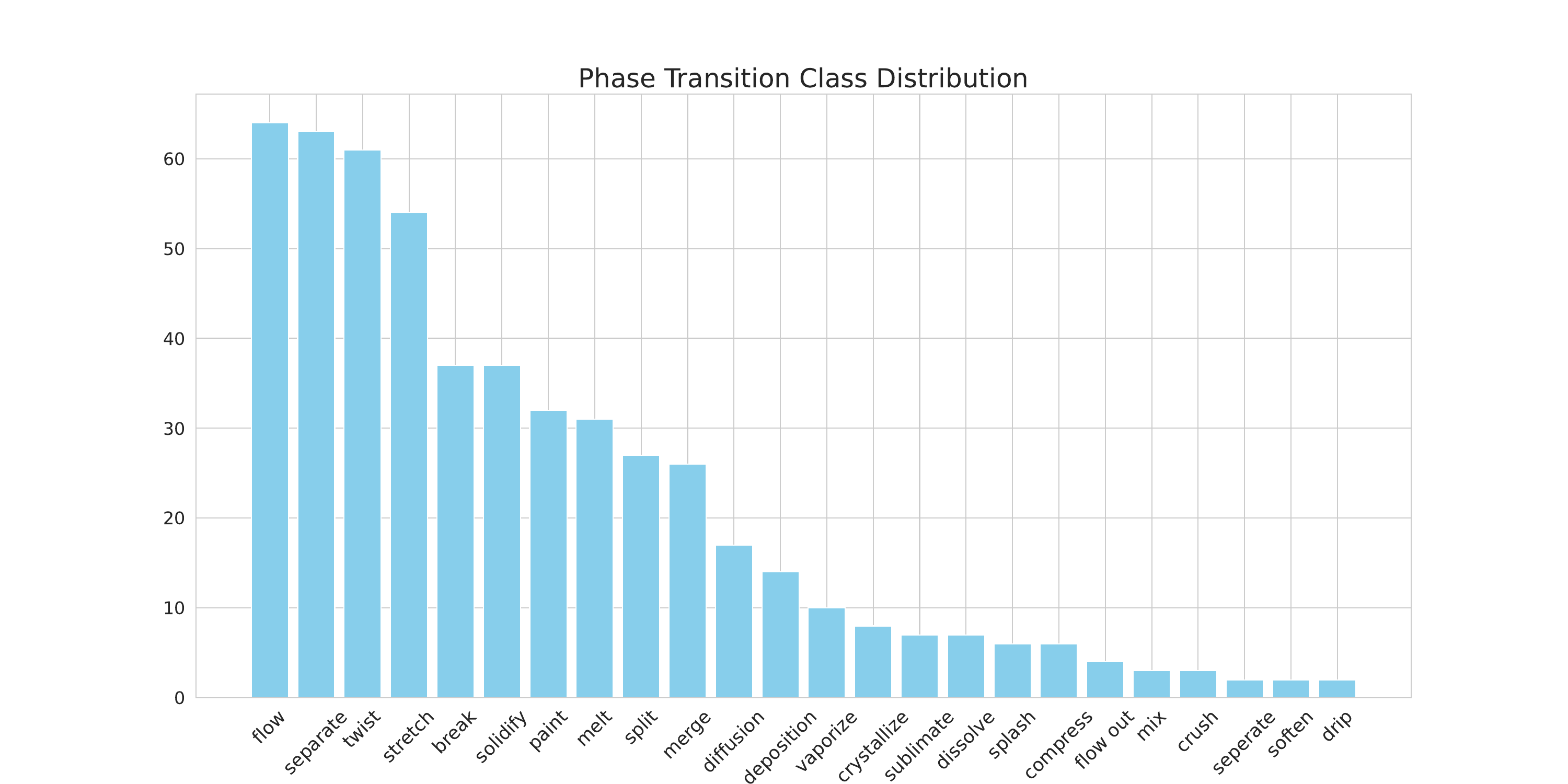} 
            \caption{Transition category.}
            \label{fig:distribution_of_transition}
        \end{subfigure}
    \end{minipage}
    \vspace{-5px}
    \caption{M$^3$-VOS characteristics. 
    (a) M$^3$-VOS dataset pipeline including annotating and dual model cross-valid method for model-bias.
    (b) Transition from before phase (left) to after phase (right).
    (c) Distribution of the phase transition in the M$^3$-VOS.}
    \vspace{-10px}
\end{figure*}

We chose to source our videos from YouTube and BiliBili mainly, where there are lots of videos about intra-phase transition and cross-phase transition in different scenery: kitchen, outdoor, factory, school, farm, \etc. 
In total, \textbf{14 scenarios} as shown in \cref{fig:dataset_pipeline}. 
In these scenarios, target objects undergo intra-phase transitions or cross-phase transitions as they engage in various interactions, including interactions with humans, other objects, and the environment.  

Most of the internet videos were captured using consumer cameras. To better capture rapid and dramatic phase transitions, such as balloon bursts or glass shattering, we also collected \textbf{high-frame-rate} video clips shot with high-speed cameras. For slower phase transitions, like ice melting or frost formation, we gathered a portion of \textbf{low-frame-rate} video clips taken with time-lapse techniques.

Next, we performed temporal cutting and spatial cropping to focus more on the process of phase transitions. The videos have an average length of 14.27 seconds.
To further enrich the scenarios in our dataset, we collected a portion of videos filmed in the laboratory. 
We have carefully checked all the videos to avoid possible ethnic problems and will only provide access through video URLs along with downloadable preprocessing scripts.

\subsection{Annotation Design } 

Each video clip is accompanied by rich annotations for target segmented objects, identifying the corresponding phase transitions. This includes a bilingual (Chinese and English) description of one or more target objects in each clip, the phases of each target object in the first and last frames, as well as pixel-level segmentation annotations with 30 fps.

In video collection, we instructed the video collectors to use clear and accurate positional terms to describe each target object in the first frame. The descriptions are ``\textit{In the process of [specific action], the [specific object] which at [specific location] in the initial frame or whose color is [specific color]}.''
, providing accurate guidance for the volunteers responsible for the mask annotations.

In the annotation, we instructed volunteers to use the information provided by the video collectors about the target objects and the semi-auto Annotation Tool (detailed in \cref{sec:Annoate_Tool}) to annotate the masks of the target objects in each frame of the video. 
For various \textit{uncertain} void masks caused by unclear image quality or motion blur, 

we instructed volunteers to annotate the corresponding void masks, similar to the approach in the VOST~\cite{tokmakov2023breaking}.
In this way, we can disregard the influence of these pixels during the evaluation process.
Volunteers were also required to select the \textit{initial} and \textit{final} phases of the target objects from \cref{table:Phase} based on their states in the video, and subsequently label the corresponding phase transformations.

\subsection{\textbf{Annotation Collection}} 

In this section, we introduce  our semi-auto Annotation tool which  efficiently annotates the masks of target objects in the videos, and our annotation collection pipeline.

\subsubsection{Semi-Auto Multi-Level Annotation Tool} 
\label{sec:Annoate_Tool}

To efficiently annotate the masks of target objects in the videos, we employed a semi-automated multi-level annotation tool that utilizes a paint-erase tool (\textbf{pixel-level}), pixel color difference assistance (\textbf{appearance-level}), and neural network-based annotation support (\textbf{object-level}).

Our annotation tool mainly employs a \textit{paint-and-erase} approach. Volunteers can use a brush and eraser tool, along with a magnification feature, to accurately annotate the masks of target objects and the void masks in the video.
To achieve high-frame-rate mask annotation, we integrated the fully automated annotation tool provided by Cutie~\cite{cheng2024putting} based on RITM~\cite{sofiiuk2021revivingiterativetrainingmask} for interactive image segmentation and Cutie for video object segmentation. After volunteers annotate masks for a specific frame using the paint-and-erase approach or point prompt method, they can utilize Cutie to propagate masks for that frame. Subsequently, the propagated pseudo-ground truth can refined using the paint-and-erase approach.
The \textit{neural network based} automatic annotation tool provides good pseudo ground truth references for objects with clear contours. However, the tool often fails to provide satisfactory pseudo-ground truth for target objects like water vapor, thick smoke, and splashes, which have unclear outlines. 
Thus, our tool integrates a \textit{color difference masking}, allowing users to select a specific pixel value in the image and set a range to create a color difference template for the video. 

Given a pixel $p_s$ and any other pixel  $p_i$, the position of $p_i$ will  be labeled according to
\begin{equation}
  \vspace{-10px}
\begin{aligned}
    M(p_i) &= 
\begin{cases}
 & \Delta H_{s}(p_i) < 0.1 \delta,\\
1  \iff  & \Delta S_{s}(p_i) < \delta,\\
  & \Delta V_{s}(p_i) < \delta, \\
0 & \text{otherwise.}
\end{cases} \\
\Delta \phi_s (p_i) &= | \phi(p_i) - \phi(p_s) |, \quad \phi \in \{H, S, V\}.
\end{aligned}
\label{equa:combined}
  \vspace{5px}
\end{equation}

We label the pixel $p_{i}$ if and only if the color difference between $p_s$ and $p_i$ is within $0.1 \delta$, and the differences in saturation and brightness are within $\delta$. 
Here, $\delta$ is a user-adjustable parameter.

Our testing has shown that this functionality effectively annotates objects with unclear contours but distinct colors, especially for gases and liquids.

\subsubsection{Annotation Pipeline and Statistics} 
\label{sec:pipeline}

As illustrated in \cref{fig:dataset_pipeline}, in our annotation pipeline, we hired 12 volunteers and provided them with 7 days of training to use the annotation tools on the videos we collected.

To ensure the quality of the data annotations, we employed 3 experienced reviewers to audit the annotated data. 
Each volunteer's masks were evaluated based on three criteria: tracking accuracy, completeness of annotation, and boundary stability, scored on a scale from 0 to 3 (detailed scoring criteria and the review result can be found in the Supplement). 
The final mask review score was the average of the two reviewers' ratings. If any score is below 2, we consider the mask Unqualified. 
In addition to the MOS reviewing, we employed a dual-model cross-validation approach to verify the annotated masks (detailed in \cref{sec:dual-Model validation}). If the validated mask annotations do not meet the quality standards, we require the volunteer to annotate again until the review is approved.

Overall, we collect \textbf{205,181} masks, with an average track duration of \textbf{14.27}s. 
M$^3$-VOS covers \textbf{120+} categories of objects across \textbf{6} phases within a total of \textbf{14} scenarios, encompassing \textbf{23} specific phase transitions.  
We report additional statistics in \cref{fig:distribution_of_transition,fig:sankey_before_after}.

\subsection{Avoidance of Model Bias} \label{sec:dual-Model validation}

Although using Cutie to generate pseudo-ground truth significantly improves the annotation efficiency, it's crucial to thoroughly verify whether the volunteers' refinements of the pseudo-ground truth can effectively mitigate the model bias introduced by Cutie. 
Therefore, we introduce our method for ensuring that the annotated masks are free from bias via dual model cross-validation. 

In this process, we sample around 20\% video clips in M$^3$-VOS and require half of our volunteers to annotate the masks using another self-implement SAM2-assist tool. Meanwhile, they are required to label masks secondly using the Cutie-assist tool. Finally, we compare three kinds of masks from the two subjective qualitative and the objective quantitative aspects. In this way, we demonstrate that our dataset pipeline effectively mitigates the model bias and that the model bias of mask annotations in M$^3$-VOS is negligible. The details of the process and result can be found.

\subsection{Core Subset}

To facilitate analysis of the challenge facing the phase transition in the VOS task, we extracted video clips that humans consider to be highly representative and comprehensive in variety. Finally, we split M$^3$-VOS into two subsets for evaluation, including the full subset and core subset. The details can be found in Supplementary.
\begin{figure}[tbp]
    \centering
    \includegraphics[width=\linewidth]{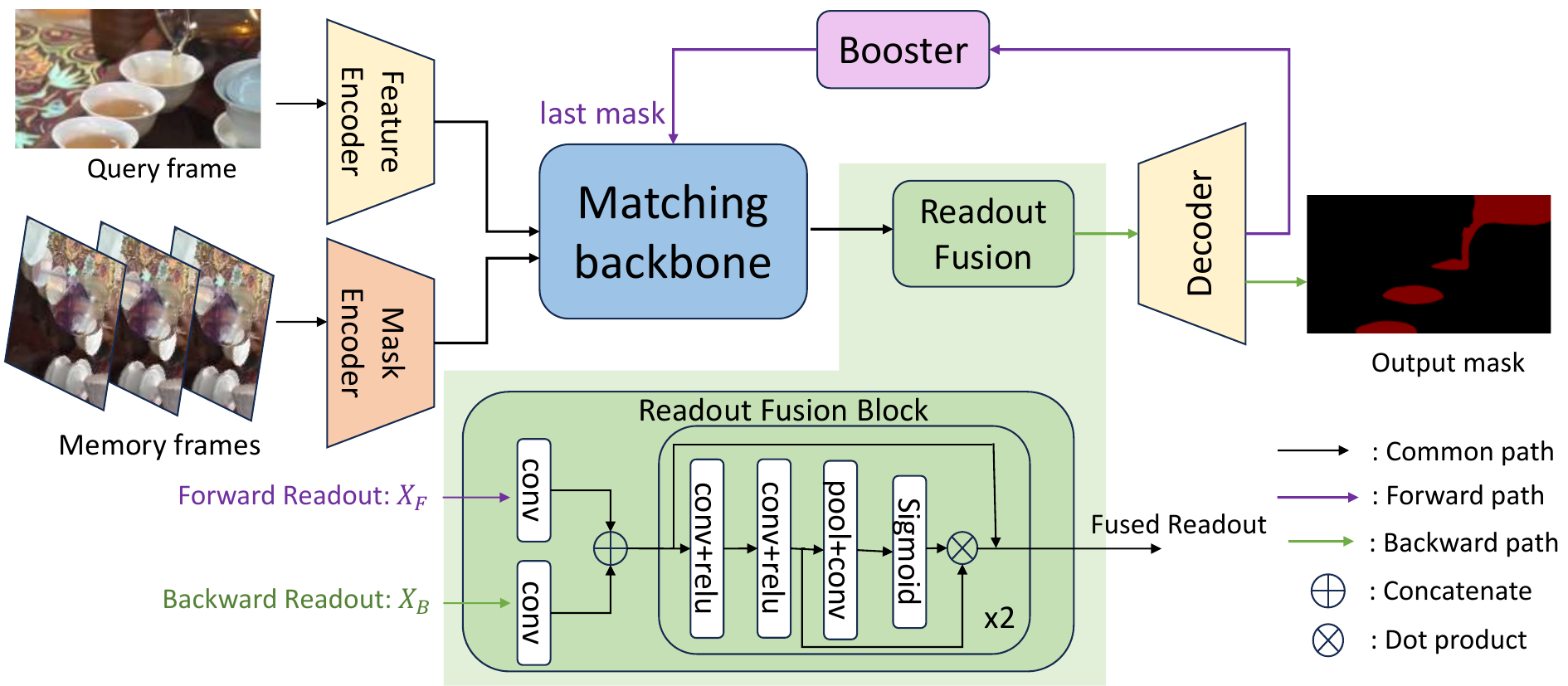}
    % \vspace{-5px}
    \caption{Overview of our ReVOS. ReVOS is a \textit{plug-and-play} based on propagation-based matching backbones. ReVOS includes a Booster Module and a Readout Fusion Module. The forward and reverse pathway is shown in purple and green separately.}
    \label{fig:algorithm}
    % \vspace{-7px}
\end{figure}

\section{Method} 
\label{sec:method}
In data collection and annotation, we observed that when objects undergo intra-phase or cross-phase transitions, their inherent \textit{disorder} tends to increase. This hidden phenomenon makes mask predictions increasingly more difficult. If the final state of the object at the end of the video is well annotated, \textit{reverse-propagating} to predict the mask becomes more accurate. 
We hypothesize that this is because the reverse process involves a gradual reduction in disorder, which simplifies segmentation. 

Given an object in a specific frame, We use Local Binary Pattern (LBP)~\cite{ojala2002multiresolution} of its mask to describe its entropy. By isolating the object from the background, we create a binary image $G$. If we calculate the LBP values of all pixels in $G$, we obtain a histogram $H$, where $H(i)$ represents the frequency of the LBP value $i$. then $h_{LBP}$ can be formulated as \cref{equ:entropy}:
\begin{equation}
    \label{equ:entropy}
     h_{LBP} = -\sum_{i=0}^{N-1} H(i) log_2H(i).
\end{equation}
We calculated the average $h_{LBP}$ for the masks in the first and second halves of several datasets. The results are shown in \cref{tab:hLBP}. In most datasets, the $h_{LBP}$ values in the latter half are higher than those in the first half, which supports our hypothesis.

Given this, we propose a paradigm to improve performance using reverse-propagation. We introduce a \textit{plug-and-play} framework named ReVOS based on mask propagation backbones (\cref{fig:algorithm}). 
We consider only the single target object as the multi-object scenarios are easily derived. First, the model performs a forward propagating pipeline based on the matching backbone, where the last output mask will be boosted and re-introduced into the model. Then the model will perform \textbf{reverse propagation} on a sliding window of previous frames. The object readout of the forward and reverse pipeline will be \textit{fused} in the Readout Fusion Module to generate better results. 
In the following, we describe the three main contributions of ReVOS.

\textbf{Reverse Memory}.
\label{subsec:reverse-memory}
In the forward propagation of the mask, we maintain a sliding window of length $T$. Suppose  the model needs to predict the $t$-th frame as the current time. The sliding window includes image information from $max(0, t-T)$ to $max(0, t-1)$ and the extracted forward memory readout.
The reverse memory is similar to working memory in \cite{cheng2022xmem}. The difference is that it only stores high-resolution features for the reverse process, serving the same function as working memory in the reverse propagation. 
It is cleared at the beginning of each reverse propagation as we want the reverse memory to only collect information from the current reverse process. 

\begin{table}[tbp]
    \centering
    \resizebox{0.97\linewidth}{!}{
    \begin{tabular}{ccccc}
    \toprule
      Dataset    & M$^3$-VOS &   VOST   &  YouTube VOS-2019 val& DAVIS' 17 val  \\
      \midrule
    First half   &  4.28   &     4.29  &     4.32  &   4.72  \\
    Latter half  &  4.37   &     4.48  &    4.35   &  4.68  \\
    \bottomrule
    \end{tabular}}
    \vspace{-5px}
    \caption{$h_{LBP}$ of the first half and the latter half of VOS datasets.}
    \label{tab:hLBP}
    \vspace{-5px}
\end{table}
\textbf{Booster}.
\label{subsec:booster}
if the model loses information about the mask for certain object parts during the forward process, this loss will be taken into the reverse process and continue to adversely affect the segmentation results. To address this issue, we implement a strategy to boost the mask during forward propagation. 
This enhancement aims to predict as many locations as possible, thereby reducing local mask loss. The boosted mask will be re-input into the matching backbone to generate the next frame mask.
In the implementation, we output the final mask via 
\begin{equation}
    M = \sigma(\alpha X_{decode}),
\end{equation}
where $\alpha$ is boosting factor, $\sigma$ is sigmoid function, $X_{decode}$ is the decoded logits, $M$ is the boosted mask.

\textbf{Readout Fusion}.
\label{subsec:readout-fusion}
In the forward and reverse process, readout features will be extracted from the matching backbone.
To obtain the final mask, we design a Fusion module to integrate readout features from the forward process, as shown in  \cref{fig:algorithm}. 

\subsection{Implementation Details}
\label{subsec:implementation-details}
We adopt Cutie-base~\cite{cheng2024putting} as our matching backbone, which is trained under ``MEGA'' setting with 5 datasets: 
YoutubeVOS~\cite{xu2018youtube}, DAVIS~\cite{perazzi2016benchmark}, BURST~\cite{athar2023burst}, OVIS~\cite{qi2022occluded}, and MOSE~\cite{ding2023mose}. In training, we froze all the parameters of Cutie and only trained the Readout Fusion Module.
We finetune our model with AdamW \cite{loshchilov2017decoupled} optimizer. The learning rate is set to 1e-5. Finetuning takes about 75k iterations and we reduce the learning rate by 10 times after 60K and 67.5K iterations. The model is trained on 4 A100 GPUs for 10 hours.

\section{Analysis}

\begin{figure}[t]
  \centering

   \includegraphics[width=\linewidth]{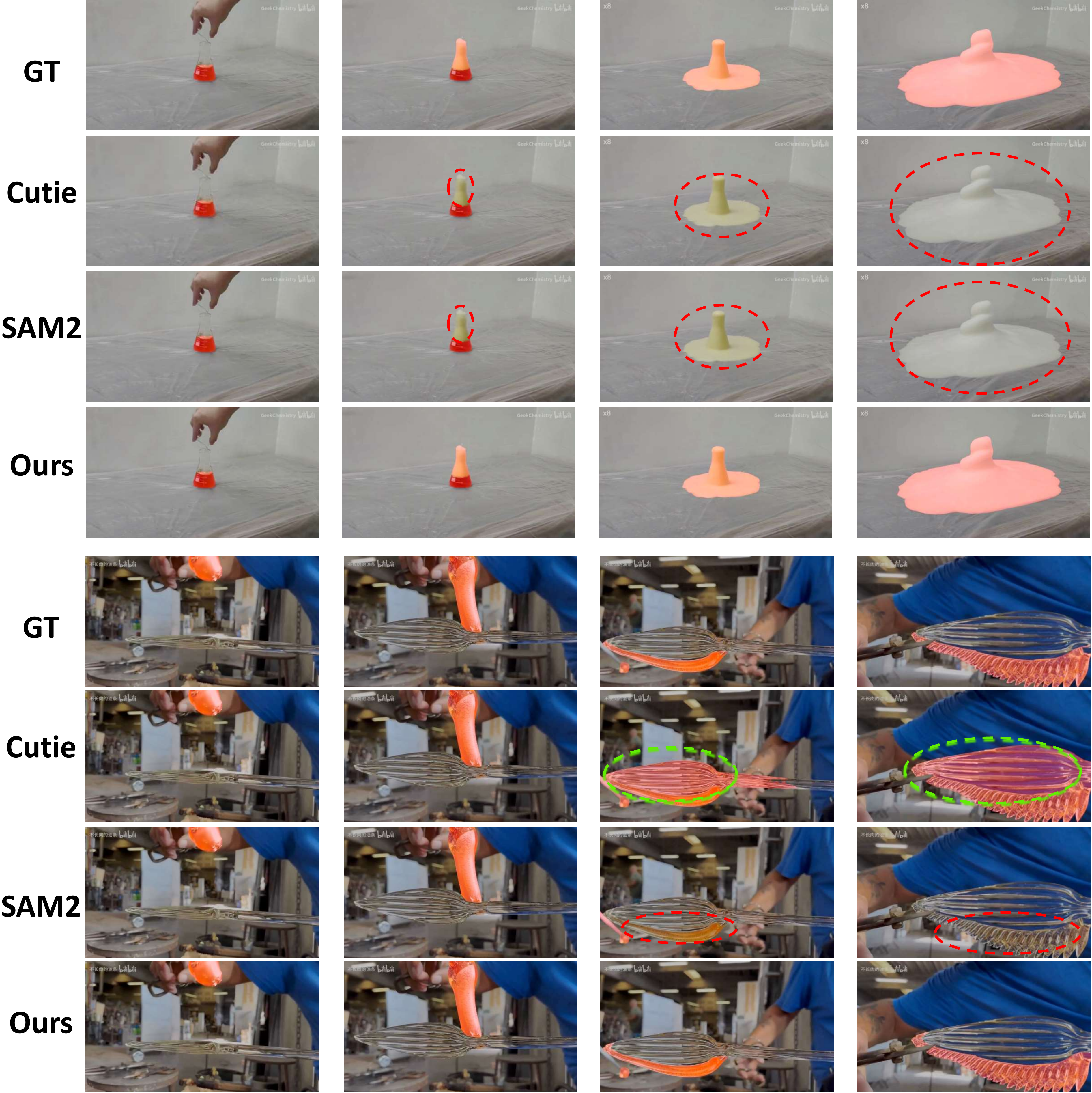}

   \caption{Visualization of qualitative results. 
   (\textcolor{red}{Red circle}: false-negative region; 
   \textcolor{green}{Green circle}: false-positive region). 
   The highlight color region is the predicted mask.
   Phase transitions, \eg, reagent eruptions and glass solidification, challenge current models in boundary prediction, causing tracking failures or background interference. However, Our reverse inference correction effectively addresses these issues.} 
   \label{fig:failure_case_half}
\end{figure}

In evaluation, we adopt standard metric Jaccard index $\mathcal{J}$, newly proposed metrics $\mathcal{J}_{tr}$~\cite{tokmakov2023breaking} and $\mathcal{J}_{cc}$ \cite{yu2023video}.  
$\mathcal{J}_{tr}$ represents the Jaccard index of the last 25\% frames and  $\mathcal{J}_{cc}$  represents the Jaccard index based on each connected component and takes an average over all components. For YouTubeVOS~\cite{xu2018youtube}, we use its official metrics: $\mathcal{J}$ and $\mathcal{F}$ for seen and 
unseen categories. $\mathcal{G}$ is the averaged $\mathcal{J\&F}$ for both seen and unseen classes.

\subsection{Main Results}

\begin{table*}[t]
    \centering
    \resizebox{\linewidth}{!}{
    \begin{tabular}{lccccccccccccccccc}
    \toprule
      & \multicolumn{3}{c}{M$^3$-VOS full}  & \multicolumn{3}{c}{M$^3$-VOS core}    & \multicolumn{3}{c}{VOST val} & \multicolumn{3}{c}{DAVIS'17 val} & \multicolumn{5}{c}{YouTubeVOS-2019 val}    \\
      \cmidrule(lr){2-4} \cmidrule(lr){5-7} \cmidrule(lr){8-10} \cmidrule(lr){11-13}    \cmidrule(lr){14-18}
       & $\mathcal{J}$ & $\mathcal{J}_{tr}$ & $\mathcal{J}_{cc}$  & $\mathcal{J}$ & $\mathcal{J}_{tr}$ & $\mathcal{J}_{cc}$  & $\mathcal{J}$ & $\mathcal{J}_{tr}$ & $\mathcal{J}_{cc}$ & $\mathcal{J}$ & $\mathcal{J}_{tr}$ & $\mathcal{J}_{cc}$    & $\mathcal{G}$ & $\mathcal{J}_{s}$ & $\mathcal{F}_{s}$  & $\mathcal{J}_u$ & $\mathcal{F}_u$ \\
      \midrule
      
     DeAOT * \cite{yang2022decoupling} & \cellcolor[HTML]{acbed2} 72.5 &\cellcolor[HTML]{acbed2} 65.2 & \cellcolor[HTML]{acbed2} 61.2 & \cellcolor[HTML]{acbed2} 62.3 & \cellcolor[HTML]{acbed2} \underline{55.1} & \cellcolor[HTML]{acbed2} 51.4 & \textbf{45.6} & \textbf{33.2} & \textbf{35.1}  & 82.7 & 79.8 & 60.9 & 86.2 & \underline{85.6}  & \textbf{90.6} & 80.0 & 88.4   \\
   
      XMem~\cite{cheng2022xmem} & 70.4  & 61.5 & 58.8 & 60.6 & 50.6 &  49.2  & 36.0 & 24.8 & 26.6 & 82.9 & 81.3 & 61.3 & 85.4  & 84.3 & 88.6 & 80.3 & 88.5 \\      
      
      RMem + DeAOT   $  ^{\dag} $~\cite{zhou2024rmem}  &  73.4 & \underline{66.1} & 62.6 & 56.1 & 45.5 & 46.7 & 40.5 & 25.3  & 31.9 & 82.3 & 79.3  & 60.9 & 85.9 & 84.6  & 89.4  & 80.8 & \underline{88.9}  \\
      
      SAM2~\cite{ravi2024sam} & 69.5 & 57.8 & 58.2 & 61.0 & 49.9 & 48.9 & \underline{44.1} & \underline{28.0} &  \underline{32.7} & 85.5 & 82.8 &  \underline{64.1} & 85.2 & 83.7  &  80.6 & \textbf{87.9} & 88.5  \\
      
      Cutie-base~\cite{cheng2024putting} & \underline{74.6}  &  64.6 & \underline{64.3}  & \underline{64.6} & 52.0 &\underline{53.9}  & 40.8 & 25.1 & 31.6  & \underline{85.6} & \underline{84.6} & 63.9  & \underline{86.8} & \textbf{85.9}  & \underline{90.4} & 81.6 & 89.3  \\
      
    \underline{Ours}   & \textbf{75.6} & \textbf{66.5} & \textbf{65.2}  & \textbf{66.3} & \textbf{55.8} & \textbf{55.5} & 41.0 & 25.3 & 31.7 & \textbf{86.0} & \textbf{84.8} & \textbf{64.2} & \textbf{86.8} & 85.3  & 89.8 & \underline{82.1} & \textbf{89.9}   \\
    \bottomrule
    \end{tabular}}
    % \vspace{-5px}
    \caption{Comparison of semi-VOS methods. 
    The metrics are $\mathcal{J}$, $\mathcal{J}_{tr}$, and $\mathcal{J}_{cc}$ for M$^3$-VOS (full, core), VOST, and DAVIS'17 ($ ^{\dag} $: R50-DeAOT-L with restricted memory bank without temporal position embedding; 
    *: SwinB-DeAOT-L; the grep region contains 469 objects. 
    The Video clip of Other Objects is too long. 
    DeAOT doesn't support long videos due to the memory limitation).}
    % \vspace{-5px}
    \label{tab:perform-main}
\end{table*}

We compare ReVOS with SOTA approaches on our dataset and standard benchmarks: VOST validation~\cite{tokmakov2023breaking} DAVIS 2017 validation~\cite{Pont-Tuset_arXiv_2017} and YouTubeVOS validation~\cite{xu2018youtubevossequencetosequencevideoobject}. For our dataset, we create two versions: M$^3$-VOS full (all cases) and M$^3$-VOS core (highly representative cases).
For a fair comparison and open-vocabulary purpose, we choose only one checkpoint that each method performs the best on the DAVIS 2017 validation set and use it to test all other datasets. For DeAOT, we excluded a few long video cases as we encountered insufficient memory during inference. 

In \cref{tab:perform-main}, ReVOS achieves better results than all SOTA methods, especially on our dataset. As we freeze the Cutie backbones during training, it is evident that the increase in our model's score relative to Cutie is entirely due to our new framework. Specifically, SAM2 achieved the best result on VOST. We assume it takes more training data similar to EPIC-kitchens~\cite{damen2018scaling} from which the VOST is constructed.

\begin{figure*}[t]
    \vspace{-5px}
    \centering
    \begin{minipage}{0.55\textwidth}
        \centering
        \resizebox{1\linewidth}{!}{
        \begin{tabular}{lcccccccccccc}
        \toprule
               &  \multicolumn{3}{c}{IS}  &    \multicolumn{3}{c}{IL}   & 
         \multicolumn{3}{c}{IG}  & \multicolumn{3}{c}{CP}   \\
         \cmidrule(lr){2-4} \cmidrule(lr){5-7} \cmidrule(lr){8-10} \cmidrule(lr){11-13} 
               & $\mathcal{J}$ & $\mathcal{J}_{tr}$& $\mathcal{J}_{cc}$&$\mathcal{J}$ & $\mathcal{J}_{tr}$ & $\mathcal{J}_{cc}$& $\mathcal{J}$ & $\mathcal{J}_{tr}$ & $\mathcal{J}_{cc}$  & $\mathcal{J}$ & $\mathcal{J}_{tr}$ & $\mathcal{J}_{cc}$ \\
         
          \midrule
  \rowcolor{gray!20}   
        DeAOT  \cite{yang2022decoupling}  & 71.6  & 64.7  & 59.8 &  72.1 & 65.0 & 59.9 & 80.0 & 74.3 & 67.9 & 73.8 & 66.0 & 65.3 \\
  
        XMem  \cite{cheng2022xmem} & 70.9  & \underline{62.8} & 58.9 & 68.2 & 58.7 & 56.1 & \underline{80.2} & \underline{71.3} & 64.7 & 74.5& 66.4 & 65.4 \\
        
        Rmem + DeAOT \cite{zhou2024rmem} & 74.5  &\textbf{67.3}  & \textbf{63.3} & 71.3  & 63.6  &  59.8 & \textbf{83.5}  & \textbf{82.0}  & \textbf{68.5} & 77.0 & \underline{70.4} & 68.3  \\
    
        SAM2 \cite{ravi2024sam} & 64.7 & 52.6 & 54.7 & 69.8 & 57.7 & 57.6 & 67.6 & 57.4 & 56.1 & 74.2 & 63.9 & 64.0 \\

        Cutie-base \cite{cheng2024putting} & 72.3 & 61.7 & 62.5 & \underline{74.1} & \underline{64.1} & \underline{63.3} & 75.8 & 64.3 & 63.6 & \underline{77.5} & 69.7 & \underline{69.0}\\

        \underline{Ours} & \textbf{73.0} & 63.2 & \underline{63.3} &\textbf{75.6} & \textbf{66.4} & \textbf{64.2} & 76.8 & 66.5 & \underline{65.4} & \textbf{78.1} &\textbf{70.8} & \textbf{69.6} \\

        \bottomrule
        \end{tabular}}
         \captionof{table}{Performance of different VOS methods in phase transitions. }
        \label{tab:perform-phases}
    \end{minipage} \hfill
    \begin{minipage}{0.42\textwidth}
        \centering
        \includegraphics[width=\linewidth]{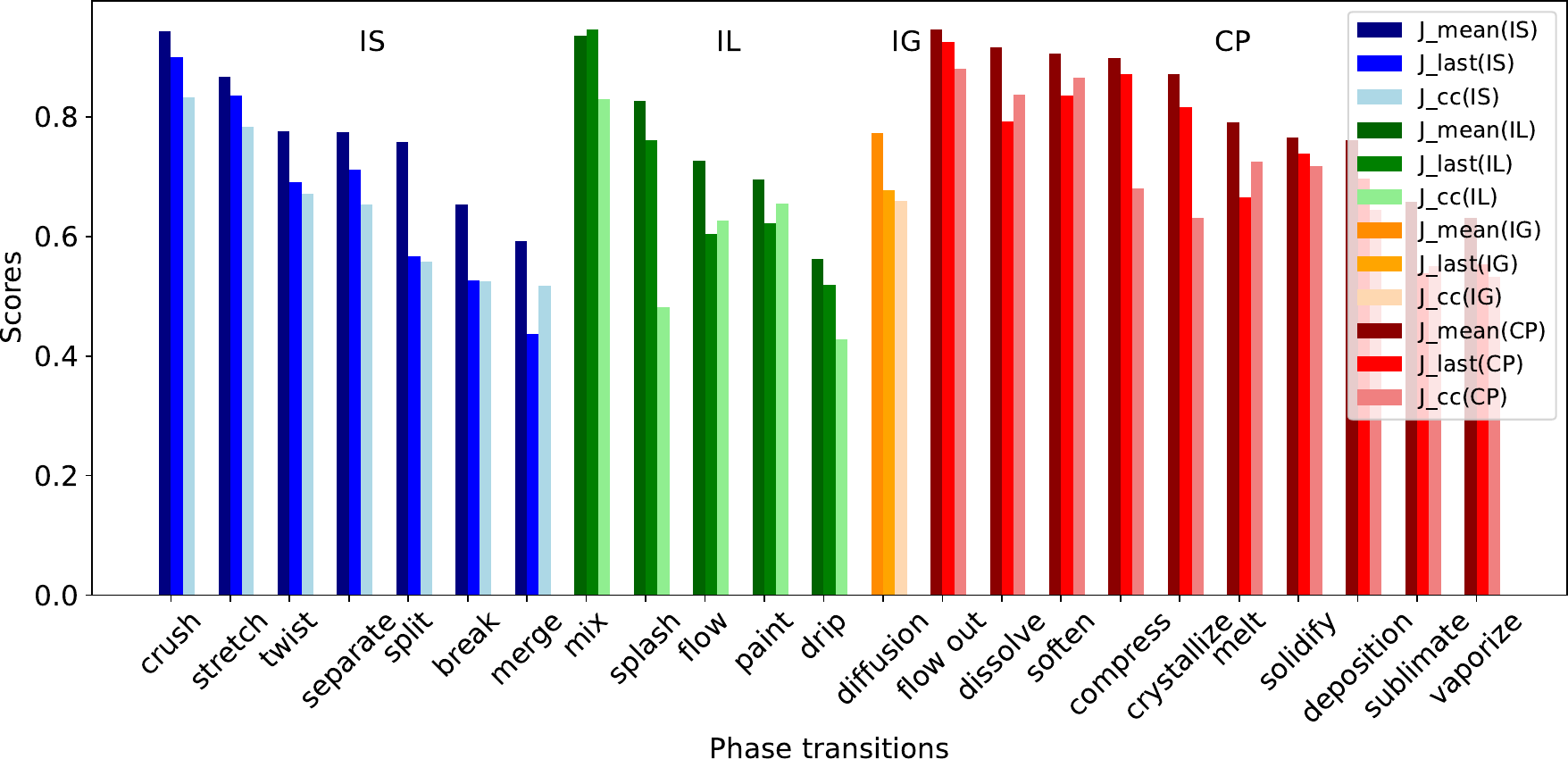}
        \vspace{-20px}
       \captionof{figure}{Results of ReVOS on phase transitions.}

        \label{fig:phase_trans}
    \end{minipage}

\end{figure*}

\subsection{Intra-Phase vs. Cross-Phase Transformations} \label{sec:phase_performance}
We present the average model performance in 4 categories: Intra-Phase (solid), Intra-Phase (Liquid), Intra-Phase (Aerosol/Gas), and Cross-Phase, which are abbreviated as IS, IL, IG and CP in \cref{tab:perform-phases}.
In addition, we also present the performance of our model for the detailed phase transitions in \cref{fig:phase_trans}.

Compared to pure Cutie, Our Cutie-ReVOS improves the performance in all phase transitions. Compared to the SOTA, Cutie-ReVOS still demonstrated excellent performance facing these phase transitions. Especially, when facing the cross-phase transition, Cutie-ReVOS achieves the best performance in all models we have evaluated.

\subsection{Performance Differences across Benchmarks}
\label{sec:main_challenge}

Compared to existing benchmarks, M$^3$-VOS significantly expands the phase range of objects, introducing a series of new challenges for tracking and segmentation. 
For example, M$^3$-VOS includes liquids and granular solids, where target objects often exhibit rapid motion due to splashing or splitting.
A large portion of fluids (both gases and liquids) also tend to be transparent, and phase transitions are often accompanied by changes in the object's appearance and color. 
As \cref{tab:perform-main} demonstrate, the performance of all existing models showed a significant decline.  

Especially, comparing  the full M$^3$-VOS full set and the M$^3$-VOS core subset, which balances the distribution of scene and phase transition, the existing modes have more obvious performance deficiencies. This phenomenon shows that the current models all have a certain degree of scenery bias and phase bias.

\subsection{Ablation Study}
We study various designs of our algorithm in the ablation.  
We report $\mathcal{J}$, $\mathcal{J}_{tr}$ and $\mathcal{J}_{cc}$ and FPS for M$^3$-VOS mid set. 

\textbf{Hyperparameter Choices}.  
\cref{tab:ablation:hyper} compares our results with different choices of hyperparameters: sliding window length $T$ and reverse interval $L$. 
Note that $T = 0$ or $L=0$ is equivalent to Cutie-base. $T$ is fixed to 30 when we vary the value of $L$ and vice versa. We find that a larger sliding window length 
means high performance but also implies a decrease in FPS.  Surprisingly, a larger reverse interval also means higher performance with an increase in FPS.
\cref{tab:ablation:booster} shows that the performance improvement brought by the booster and readout fusion demonstrates the necessity of adding these modules.

\begin{table}[t]
    \centering
    \vspace{-5px}
    \begin{minipage}{0.4\linewidth}
        \centering
        \resizebox{\linewidth}{!}{
        \begin{tabular}{ccccc}
            \toprule
            Setting  & $\mathcal{J}$ & $\mathcal{J}_{tr}$ & $\mathcal{J}_{cc}$ & FPS\\
            \midrule
            \multicolumn{5}{l}{\textit{Sliding window length}} \\
            \midrule
            10 & 75.6 & 66.6 & 65.2 & 15.3\\
            30 & \textbf{75.7} & \textbf{66.4} & \textbf{65.3} & \textbf{15.4} \\
            60 & 75.4 & 66.3 & 65.0 & 10.4 \\
            \bottomrule
            \multicolumn{5}{l}{\textit{Reverse interval}} \\
            \midrule
            10  & 75.5 & 66.5 & 65.1 & 7.9 \\
            30 & \textbf{75.7} & 66.4 & 65.3 & 15.4 \\
            60 & 75.7 & \textbf{66.7} & \textbf{65.4} & \textbf{21.6} \\
            \bottomrule
        \end{tabular}}
            % \vspace{-5px}
        \caption{Comparison with different hyperparameters.}
        \label{tab:ablation:hyper}
            \vspace{-5px}
    \end{minipage}%
    \hfill
        \vspace{-5px}
    \begin{minipage}{0.55\linewidth}
        \centering
        \resizebox{\linewidth}{!}{
        \begin{tabular}{cccc}
            \toprule
            Setting  & $\mathcal{J}$ & $\mathcal{J}_{tr}$ & $\mathcal{J}_{cc}$ \\
            \midrule
            Only forward (Cutie) & 74.7 & 64.6 & 64.5 \\
            Only backward & 75.4 & 66.4 & 64.9 \\
            With readout fusion & \textbf{75.7} & \textbf{66.4} & \textbf{65.3} \\
            \midrule
            Without booster & 74.2 & 64.2 & 63.9 \\
            With booster & \textbf{75.7} & \textbf{66.4} & \textbf{65.3} \\

            \bottomrule
        \end{tabular}}
        % \vspace{-5px}
        \caption{Ablation on Booster and Readout Fusion Module.}
        \label{tab:ablation:fusion}

        \label{tab:ablation:booster}
    \end{minipage}
    \vspace{-5px}
\end{table}

\section{Discussion and Limitations}
\label{sec:formatting}

\textbf{Annotation Bias}.  
Annotation bias is inevitably introduced by annotators' tendencies, inherent biases in assisted tools, and ambiguous regions in videos. As discussed in \cref{sec:dual-Model validation}, despite our implementation of multi-round, multi-level review processes, some discrepancies remain across annotation instances. Establishing a more standardized and rigorous data annotation review workflow could effectively mitigate annotation bias.

\textbf{Model Performance}. 
All models show limitations on our new benchmark with challenging object phase transitions.
In future works, models may be advanced by leveraging the physical knowledge embedded in MLLMs.

\section{Conclusion}
In this work, we explored the ability of the visual models to understand object phase transitions through a video segmentation task, introducing a fine-grained text-visual annotated open-vocabulary benchmark  M$^3$-VOS.
It includes 14 scenes, encompassing 120+ objects across 6 phase states and 23 phase transitions (13 intra-phases, 10 inter-phases). 
To address the limitations of existing methods, we proposed a plug-and-play module ReVOS based on entropy theory, achieving decent improvements in  M$^3$-VOS and comparable performance on other benchmarks. 
However, effectively understanding phase transitions remains a challenge. 
We believe this will pave the way for a new direction in research focused on a deeper understanding of objects.
%-------------------------------------------------------------------------

{
    \small
    \bibliographystyle{ieeenat_fullname}
    \bibliography{main}
}

\clearpage
\setcounter{page}{1}
\maketitlesupplementary

\section*{Overview}

We introduce:

 $\bullet$ More implementation details of our work in \cref{sec:detail_annotate,sec:detail_metric,sec:SQA,sec:Model_Bias,sec:details_tool,sec:Core}.

 $\bullet$  More experiments about the challenge in M$^3$-VOS in \cref{sec:challenge}.

 $\bullet$  More failure cases in \cref{sec:failure} .

\section{Details of Annotations} \label{sec:detail_annotate}

\subsection{Phase Definition}

We list the specific definitions of phase below:

\begin{itemize}
\vspace{-1px}
    \item \textbf{Solid}: Volume is relatively fixed, has distinct boundaries, and shapes independent of the container.
    \begin{itemize}
        \item \textbf{Particulate:} Composed of several fragmented parts.
        \item \textbf{Non-particulate:}  Composed of single/few larger parts. 
        \begin{itemize}
            \item \textbf{Rigid Body:} Exhibiting a relatively fixed shape and resistance to deformation.
            \item \textbf{Flexible Body:} Has a relatively unstable shape and can undergo deformation easily.
        \end{itemize}
    \end{itemize}
    
    \item \textbf{Liquid:}  Volume is relatively fixed and has 
    distinct boundaries, fluidity, or shape dependent on the container.
    \begin{itemize}      
        \item \textbf{Viscous Fluid:} 
            Has significant viscosity, and can stretch.        
        \item \textbf{Non-viscous Liquid: }
            No significant viscosity, cannot stretch.    
    \end{itemize}
    
    \item \textbf{Aerosol/Gas:} Volume not fixed, has no distinct boundaries, shape dependent on the container.
   
\end{itemize}

\subsection{Phase Transition Definition}

In our works, we ensure that the definition of phase transitions meets a fundamental requirement:
the transition from an initial state to a final state may correspond to different specific phase transitions depending on the characteristics of the transformation.
However, for a specific phase transition, its initial and final states must be unique.

We list all the initial and final states for each phase transition as \cref{table:phase_transition}. Besides, in \cref{tab:define_transition}, we give a detailed definition of different phase transitions.

\begin{table*}[]
\caption{The different phase transitions and the unique initial state and final state. We give some examples to highlight their characteristic.}\label{table:phase_transition}
    \renewcommand{\arraystretch}{1.2}  
    \setlength{\tabcolsep}{1.2mm}{

\begin{tabular}{l c c c c}
\toprule
\textbf{Category}                                                                     & \textbf{Phase Transition} & \textbf{Initial State} & \textbf{Final State}&\textbf{Example} \\ 
\midrule
% \hline
\multirow{7}{*}{\begin{tabular}[c]{@{}l@{}}\textbf{Intra-Phase}\\ \textbf{(Solid)}\end{tabular}}  & Separate         & Rigid         & Rigid       &\textit{ Disassembling a gun}, \textit{Taking apart building blocks} \\
                                                                                & Twist         & Flexible      & Flexible    &  \textit{Knead dough}, \textit{Tie shoelaces}           \\ 
                                                                      & Break            & Rigid         & Particulate & \textit{Break cups}, \textit{Chop vegetables}           \\
                                                                                & Stretch          & Flexible      & Flexible    & \textit{Pull noodles},
                                                                                \textit{Pull rubber}\\ 
                                                                                & Split            & Particulate   & Particulate & \textit{Sift the rice}, \textit{Sieve the sand}           \\
                                                                                & Merge            & Rigid         & Rigid       &   \textit{assemble guns}, \textit{Jigsaw puzzle}         \\ 
                                                                                & Crush            & Rigid         & Particulate    &     \textit{Grind the herb}, \textit{Crush the stone}       \\
                                                                                \midrule
\multirow{5}{*}{\begin{tabular}[c]{@{}l@{}}\textbf{Intra-Phase}\\ \textbf{(Liquid)}\end{tabular}} & Flow             & Non-Viscous   & Non-Viscous &     \textit{Pour Water}, \textit{Pour tea}             \\ 
                                                                                & Paint            & Liquid        & Liquid      &    \textit{Paint the wall}, \textit{Paint in oil}     \\ 
                                                                                & Splash           & Non-Viscous   & Non-Viscous &  \textit{Diving sports}, 
                                                                                \textit{ Cast a stone into the water}\\ 
                                                                                & Mix              & Non-Viscous   & Non-Viscous &   \textit{Milk pouring art},
                                                                                \textit{Paint mixing with water }\\ 
                                                                                & Drip             & Non-Viscous   & Non-Viscous & \textit{ Drip the acid},\textit{ Drip the eye drops  }         \\ 
                                                                                \midrule
\begin{tabular}[c]{@{}l@{}}\textbf{Intra-Phase}\\ \textbf{(Aerosol/Gas)}\end{tabular}             & Diffuse          & Aerosol/Gas   & Aerosol/Gas &    \textit{Smoke spreads}, \textit{Mist spreads}        \\ 
\midrule
% \hline
\multirow{10}{*}{\textbf{Cross-Phase}}                                                   & Solidify          & Liquid        & Solid       &  \textit{Water freezes}, \textit{Chocolate hardens into solid chocolate}          \\
                                                                                & Melt             & Solid         & Liquid      &    \textit{Melt chocolate}, \textit{Melt the ice}        \\ 
                                                                                & Deposition       & Aerosol/Gas   & Solid       &   \textit{Form dew}, \textit{Condense into alcohol}    \\ 
                                                                                & Vaporize         & Liquid        & Aerosol/Gas & \textit{Humidifier sprays water}, \textit{Boil water}               \\ 
                                                                                & Crystallize      & Liquid        & Solid       & \textit{Making salt}, \textit{Making sugar   }       \\
                                                                                & Sublimate        & Solid         & Aerosol/Gas & \textit{Burn coal}, \textit{Burn plastic}           \\
                                                                                & Dissolve         & Solid         & Liquid      &  \textit{Dissolve the tablet}, \textit{Make formula}          \\
                                                                                & Compress         & Solid         & Liquid      & \textit{Juicing fruits}, \textit{Extracting pomegranate juice}         \\  
                                                                                & Flow out         & Solid         & Non-Viscous &    \textit{Break chocolate with a liquid center}        \\ 
                                                                                & Soften           & Solid         & Viscous     & \textit{Boil sugar}, \textit{Bake cheese}           \\ 
                                                                                % \hline
                                                                                \bottomrule
\end{tabular}

}
    \end{table*}

\begin{table*}[]
\caption{The detail of the definition of different phase transitions.}\label{tab:define_transition}
    \renewcommand{\arraystretch}{1.2}  
    \setlength{\tabcolsep}{1.2mm}{
    
\begin{tabular}{l c c }
\toprule
\textbf{Category}                                                                     & \textbf{Phase Transition} &\textbf{Definition} \\ 
\midrule
% \hline
\multirow{7}{*}{\begin{tabular}[c]{@{}l@{}}\textbf{Intra-Phase}\\ \textbf{(Solid)}\end{tabular}}  & Separate       &  Block-like solid objects are disassembled into multiple block-like pieces  \\
                                                                                & Twist           &  Flexible objects are deformed into various shapes.           \\ 
                                                                      & Break         & Solid objects are shattered into countless small fragments.   \\
                                                                                & Stretch           & Flexible objects are elongated into a longer form.\\

                                                                             & Split            &      The solid particles disperse in all directions, spreading out from the source. 
                                                                                \\
                                                                                & Merge               &   Multiple block-like objects are combined into a single whole.        \\ 
                                                                                & Crush             &     Solid block-like objects are ground into powdery granules.       \\
                                                                                \midrule
\multirow{5}{*}{\begin{tabular}[c]{@{}l@{}}\textbf{Intra-Phase}\\ \textbf{(Liquid)}\end{tabular}} & Flow          &    The liquid moves as a whole under the influence of external force.          \\ 
                                                                                & Paint             &    The liquid is applied onto a solid surface.  \\ 
                                                                                & Splash          &   The liquid is scattered in all directions due to a sudden external force.
                                                                                \\ 
                                                                                & Mix             &   
                                                                       One liquid is poured into another, causing the two liquids to blend together.         \\ 
                                                                                & Drip            & A small amount of liquid is transferred drop by drop.        \\ 
                                                                                \midrule
\begin{tabular}[c]{@{}l@{}}\textbf{Intra-Phase}\\ \textbf{(Aerosol/Gas)}\end{tabular}             & Diffuse           &     The gas or aerosol spreads out, gradually expanding its presence in the air.      \\ 
\midrule
% \hline
\multirow{10}{*}{\textbf{Cross-Phase}}                                                   & Solidify          &  The liquid turns into a solid as it cools or hardens.        \\
                                                                                & Melt                 &    The solid turns into a highly fluid liquid.      \\ 
                                                                                & Deposition           &   The gas directly transforms into a solid without passing through the liquid state.
                                                                                \\ 
                                                                                & Vaporize          & The liquid turns into a gas as it heats up and evaporates.              \\ 
                                                                                & Crystallize           & 
                                                                                The solid crystals form and separate out from the liquid. 
                                                                                \\
                                                                                & Sublimate      & The solid directly produces gas as it transitions without becoming a liquid.        \\
                                                                                & Dissolve             &  The substance disperses evenly in the liquid, forming a solution.       \\
                                                                                & Compress             &  The solid is squeezed under pressure, forcing a large amount of liquid to be released.         \\  
                                                                                & Flow out         &    The liquid content flows out from within the solid as it is released or displaced.        \\ 
                                                                                & Soften               & The solid gradually turns into a thick, viscous liquid.           \\ 
                                                                                % \hline
                                                                                \bottomrule
\end{tabular}

}
\end{table*}

\section{Connected Component Jaccard Index}   
\label{sec:detail_metric}

To avoid ignorance of the small part during evaluation, we introduce the connect component  Jaccard Index $\mathcal{J}_{cc}$. The definition of $\mathcal{J}_{cc}$ is the average Jaccard Index of the maximum bipartite matching corresponding to all connected mask components between the ground truth and the predicted image.

We implemented our $\mathcal{J}_{cc}$ using the \textit{Hungarian} algorithm, different from the one in the official implementation of VSCOS~\cite{{yu2023video}} that calculates the $\mathcal{J}_{cc}$ using a two-loop matching process, \ie, iteratively finding for each connected component in Mask A the one in Mask B that maximizes the Jaccard Index.

\section{Details of Masks SQA}
\label{sec:SQA}

\subsection{Three Criteria in Masks SQA}
\label{sec:critera}

We design three  criteria  to evaluate the annotation in M$^3$-VOS, including:

\begin{itemize}
    \item \textbf{Tracking Accuracy}  
    \begin{itemize}
        \item 0: Target is lost or tracked incorrectly for a long time.  
        \item 1: Target is lost or tracked incorrectly for a short continuous period.  
        \item 2: Target is lost or tracked incorrectly in a few isolated frames.  
        \item \textbf{3: Target is always tracked correctly.}  
    \end{itemize}
    \item \textbf{Mask Annotation Completeness}  
    \begin{itemize}
        \item 0: Mask has been completely missing for a long time.  
        \item 1: Mask has been partially missing for a long time.  
        \item 2: Mask is partially missing in some frames.  
        \item \textbf{3: Mask is complete and accurate throughout.}  
    \end{itemize}
    \item \textbf{Mask Boundary Stability}  
    \begin{itemize}
        \item 0: Mask boundary shows an obvious jitter for a long time.  
        \item 1: Mask boundary shows a slight jitter for a long time.  
        \item 2: Mask boundary shows a slight jitter for a short time.  
        \item \textbf{3: Mask boundary shows no visible jitter.}  
    \end{itemize}
\end{itemize}

\subsection{SQA Analyze}

\begin{figure}[tbp]
    \centering
    \includegraphics[width=\linewidth]{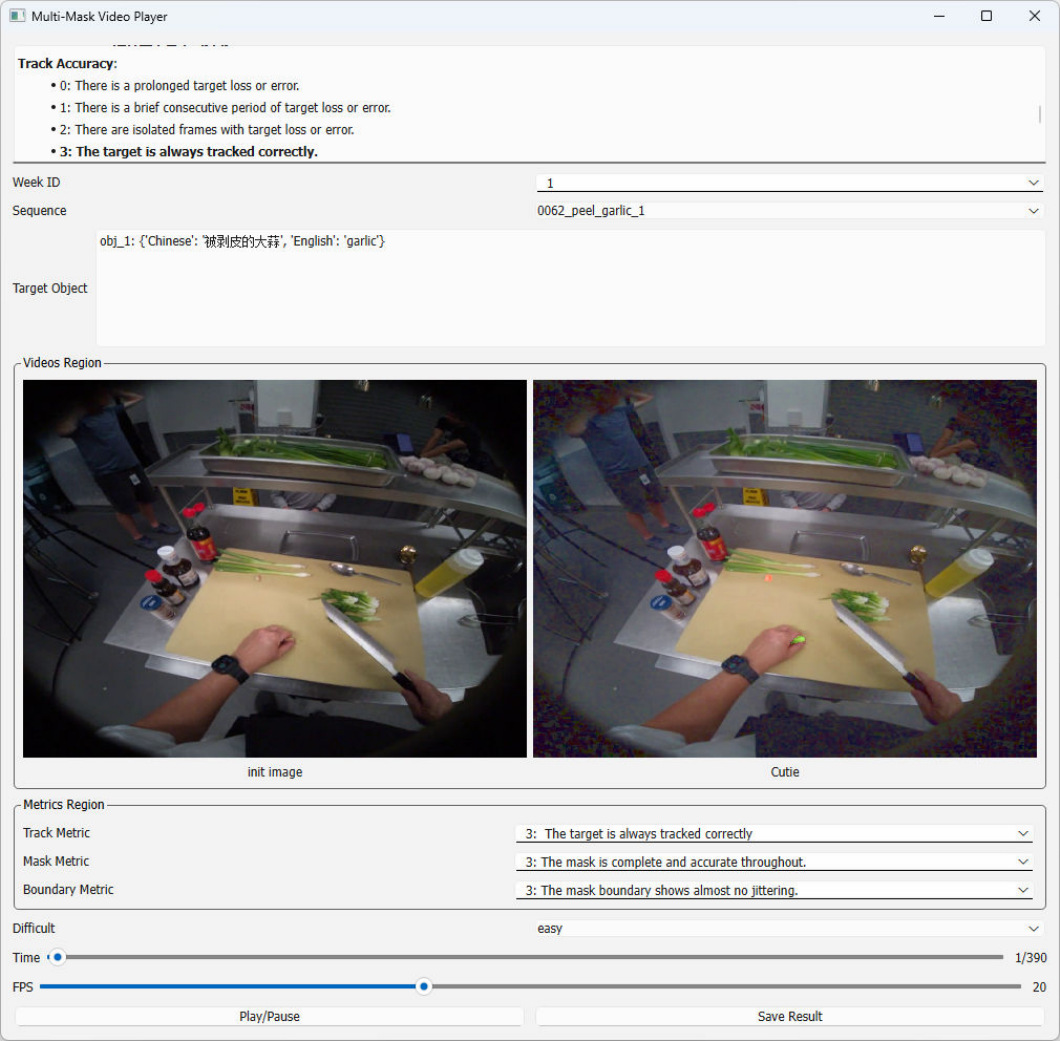}
    \caption{Review UI. The reviewer is required to evaluate the mask annotation from three  criteria. } 
    \label{fig:review_UI}
\end{figure}

We select three experienced reviewers to evaluate all of our masks in M$^3$-VOS in the criteria of \cref{sec:critera} using the reviewer UI as \cref{fig:review_UI}. In the process of constructing M$^3$-VOS, we make sure the scores in any criterion of all of our masks are higher than 2. In the final evaluation of M$^3$-VOS, the MOS in these criteria are 2.95 in tracking accuracy, 2.91 in mask annotation completeness, and 2.89 in mask boundary stability.

\section{Details of Avoidance of Model Bias}  \label{sec:Model_Bias}

In this part, we introduce the details of the dual-model cross-validation method. In this process, we validate that our dataset pipeline efficiently declines the model bias of annotations.

\subsection{IoU Analysis } 
\label{sec:IoU_ana}

In terms of model selection of the dual-model cross-validation process, we adapt the annotation model to the latest SAM2~\cite{ravi2024sam}. We utilized the open-source base plus model configuration and checkpoints, as this configuration is more effective in fully segmentation~\cite{ravi2024sam,instaboost,srda} of our target objects compared to other model setups.

In the dual-model cross-validation, we first randomly sampled a subset of videos annotated by Cutie at a ratio of 5:1. We selected 6 volunteers from a total of 12 to re-annotate this subset using both the SAM2-assisted and Cutie-assisted annotation tools, resulting in masks designated as Mask A and Mask B, respectively. To balance annotation efficiency and validation effectiveness, we set the annotation frame rate to 6 fps in the cross-validation. The high-frame-rate annotated masks obtained for the dataset are referred to as Mask O.

\begin{figure}[tbp]
    \centering
    \begin{subfigure}[b]{0.32\linewidth} 
        \centering
        \includegraphics[width=\linewidth]{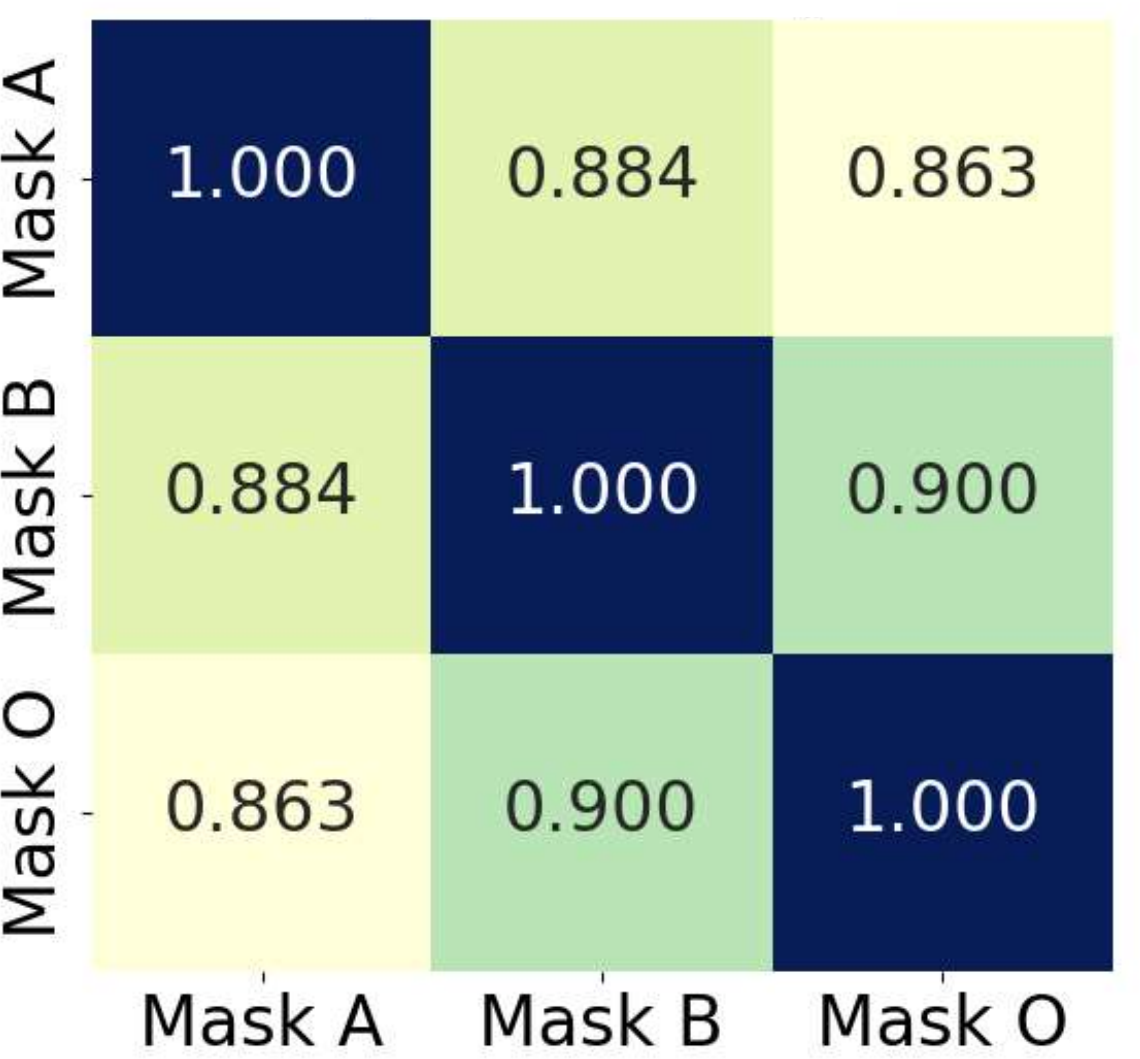}
        \caption{$J_{mean}$ Hot map.}
        \label{fig:hot_map_a}
    \end{subfigure}
    \hfill
    \begin{subfigure}[b]{0.32\linewidth}
        \centering
        \includegraphics[width=\linewidth]{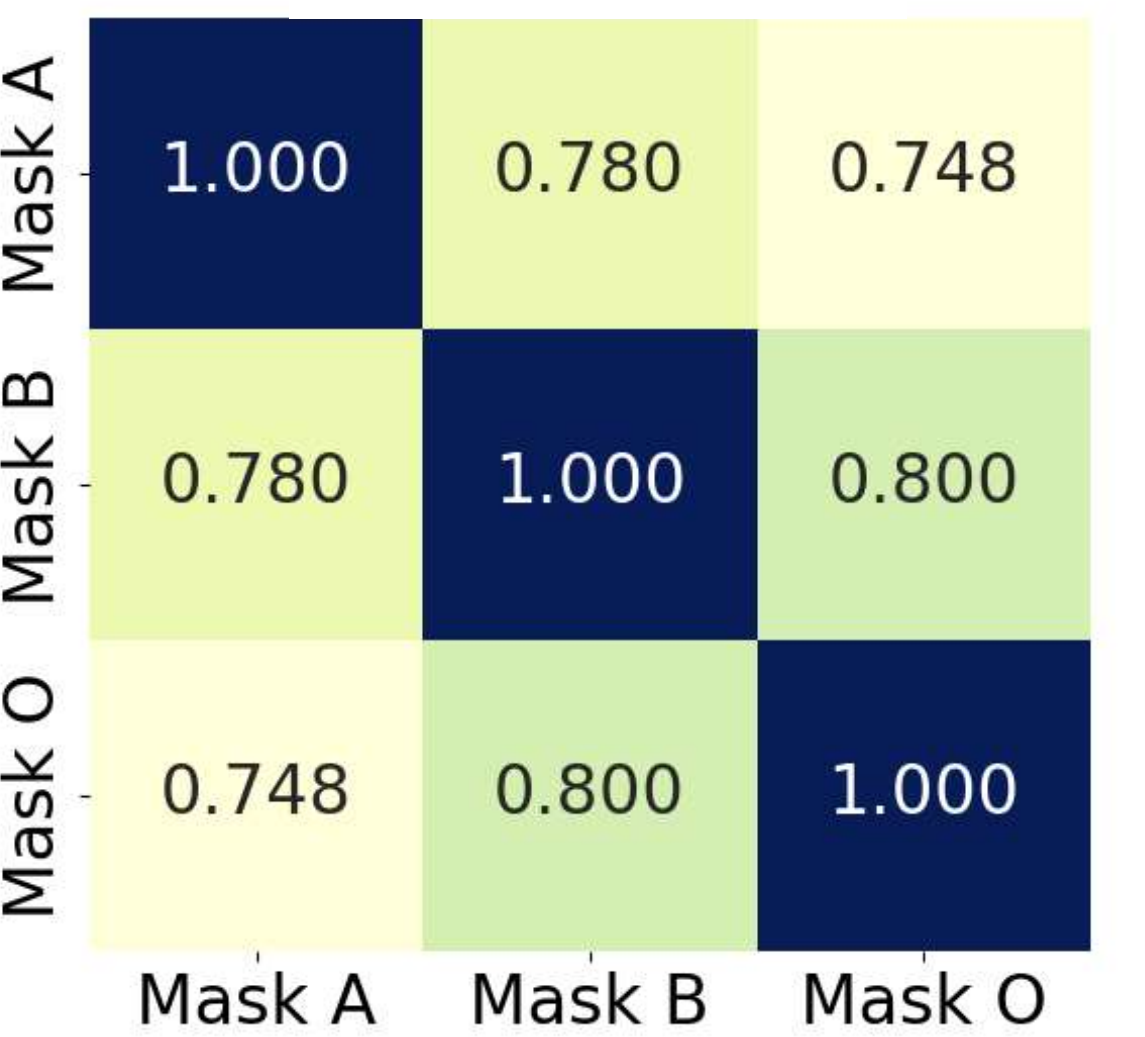}
        \caption{$J_{cc}$ Hot map.}
        \label{fig:hot_map_b}
    \end{subfigure}
    \hfill
    \begin{subfigure}[b]{0.32\linewidth}
        \centering
        \includegraphics[width=\linewidth]{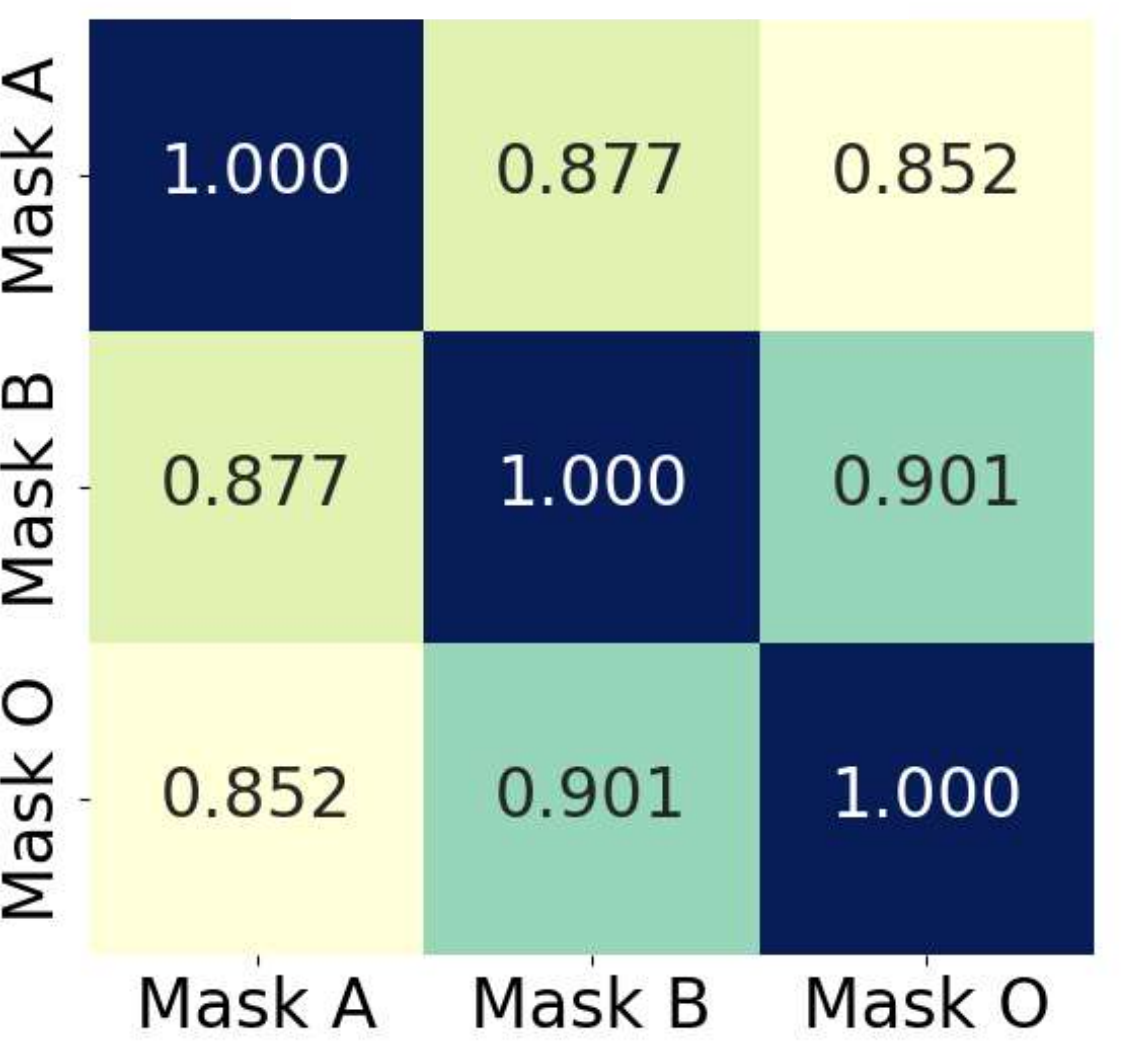}
        \caption{$J_{st}$ Hot map.}
        \label{fig:c}
    \end{subfigure}
    \vspace{-5px}
    \caption{IOU Analysis among Mask A (SAM2), Mask B (Cutie Dual), and Mask O (Cutie in the final dataset).}
    \vspace{-5px}
    \label{fig:IOU_analyze_hot_map}
\end{figure}

By calculating the Intersection over Union (IoU) and the other two metrics introduced in \cite{yu2023video,tokmakov2023breaking}, the results are shown in \cref{fig:IOU_analyze_hot_map}, indicated that $J_{st}$ (MaskA, MaskB) and $J_{mean}$   exceeded 85\% and were very close to each other.
Specifically, although the difference between Mask A and Mask B is slightly larger than that between Mask B and Mask O, we have \cref{equ:ana} holds:  
\begin{equation}  
    \label{equ:ana}
     \mathcal{J}_{\sigma}(B, O) -   \mathcal{J}_{\sigma}(A, O) \ll 1 -  \mathcal{J}_{\sigma}(B, O).
\end{equation} 

These results suggest that the annotations SAM2-assisted annotation tool produced are comparable to those of the Cutie-assisted tool, without significant bias due to model differences. 
They also indicate that the bias introduced by the models can be considered negligible compared to other sources of systematic error, such as volunteer annotation habits and inadvertent jitters during annotation.

\subsection{Blind Review: DMOS Evaluation}

\begin{figure}[t]
  \centering
   \includegraphics[width=0.98\linewidth]{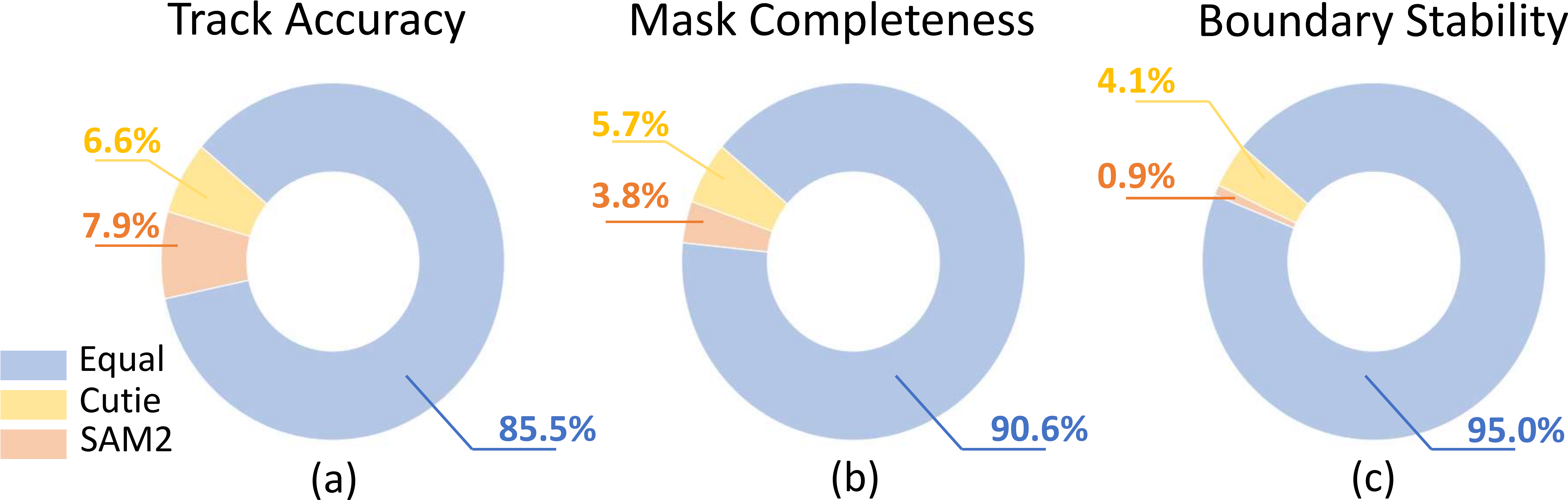} 
   \vspace{-5px}
   \caption{The result of blind review. Through subjective evaluation of three metrics (a) Track Accuracy, (b) Mask Completeness, and (c) Boundary Stability, we found that the Mask results obtained with Cutie-assisted annotation and SAM2-assisted annotation show little difference in performance. }
   \label{fig:Blind_res}
   \vspace{-5px}
\end{figure}

In addition to the quantitative analysis of model bias conducted using the dual-model validation, we also designed a mechanism for blind comparison by experienced reviewers. 
We presented 3 reviewers with both Mask A and Mask B from the cross-validation annotation process, allowing them to evaluate the performance of the two masks based on the three criteria mentioned in~\cref{sec:critera}. The reviewers were instructed to select the mask they deemed superior. If they considered the performances to be equivalent, they could choose \textit{Equal}. 
Throughout this process, the order of Mask A and Mask B was randomized to ensure that the reviewers were unaware of which mask corresponded to which model.

The final subjective evaluation results are shown in \cref{fig:Blind_res}, indicating that the two masks demonstrated a considerable degree of consistency across the three subjective evaluation metrics, with no significant bias observed.

\section{Multi-Level Semi-Auto Annotation Tool} 
\label{sec:details_tool}

\begin{figure*}[tbp]
    \centering
    \begin{subfigure}[b]{0.48\linewidth}  
        \centering
        \includegraphics[width=\linewidth]{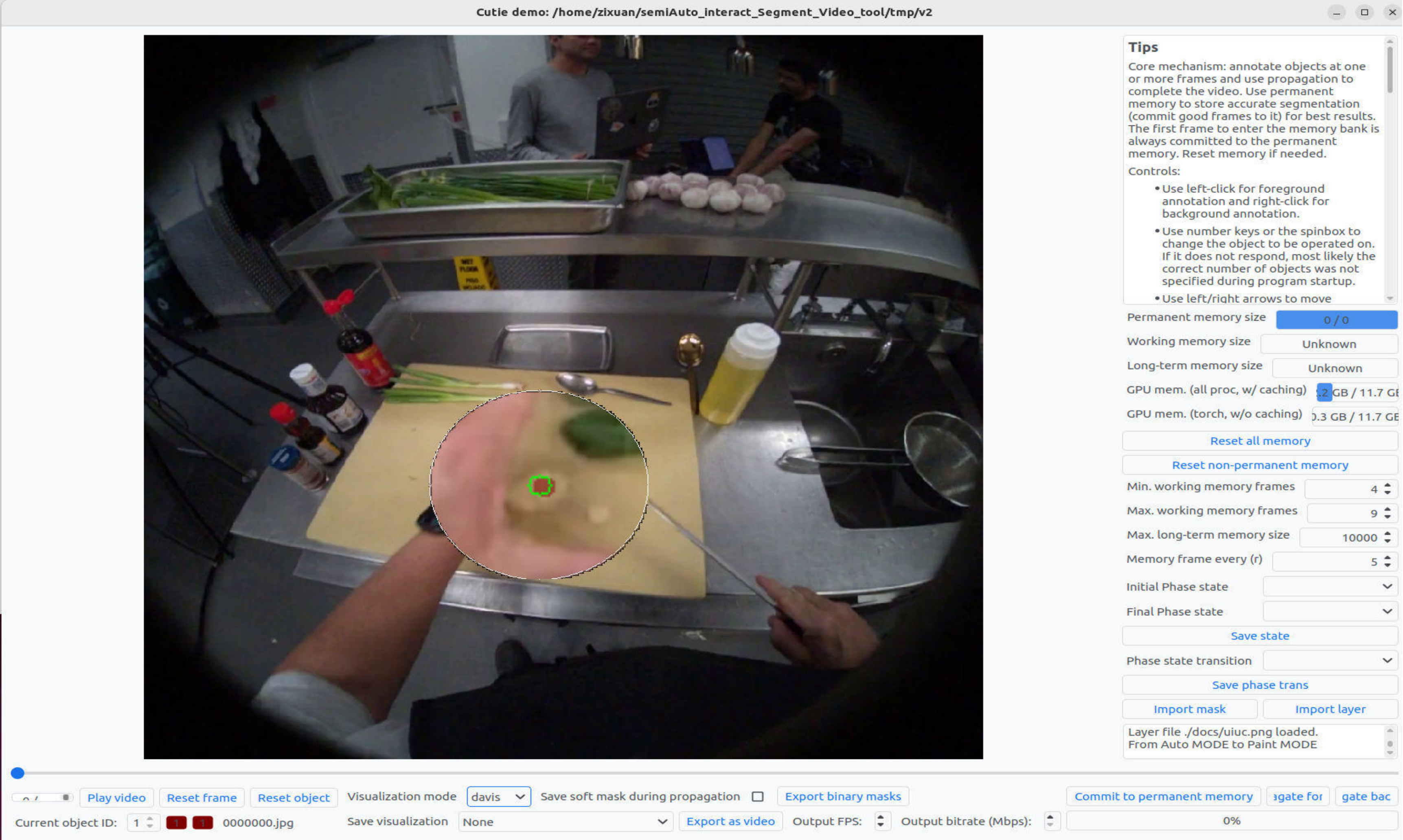}
        \caption{ \textbf{Pixel level}: brush with magnifying glass.}
        \label{fig:pixel_level_a}
    \end{subfigure}
    \hfill
    \begin{subfigure}[b]{0.48\linewidth}
        \centering
        \includegraphics[width=\linewidth]{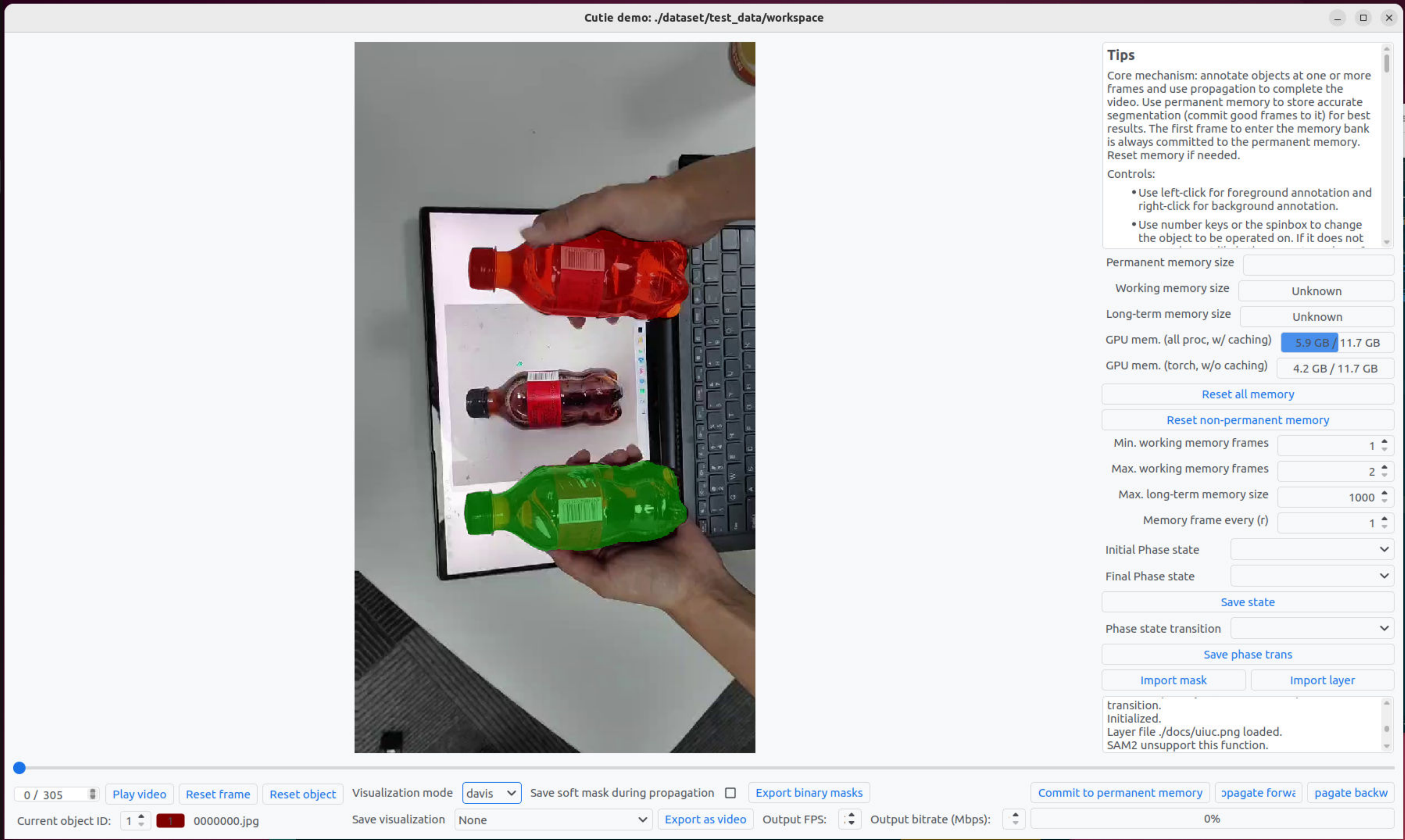}
        \caption{ \textbf{Object level}: apply click prompt to mask object.}
        \label{fig:object_level_b}
    \end{subfigure}

    \centering
    \begin{subfigure}[b]{0.48\linewidth}  
        \centering
        \includegraphics[width=\linewidth]{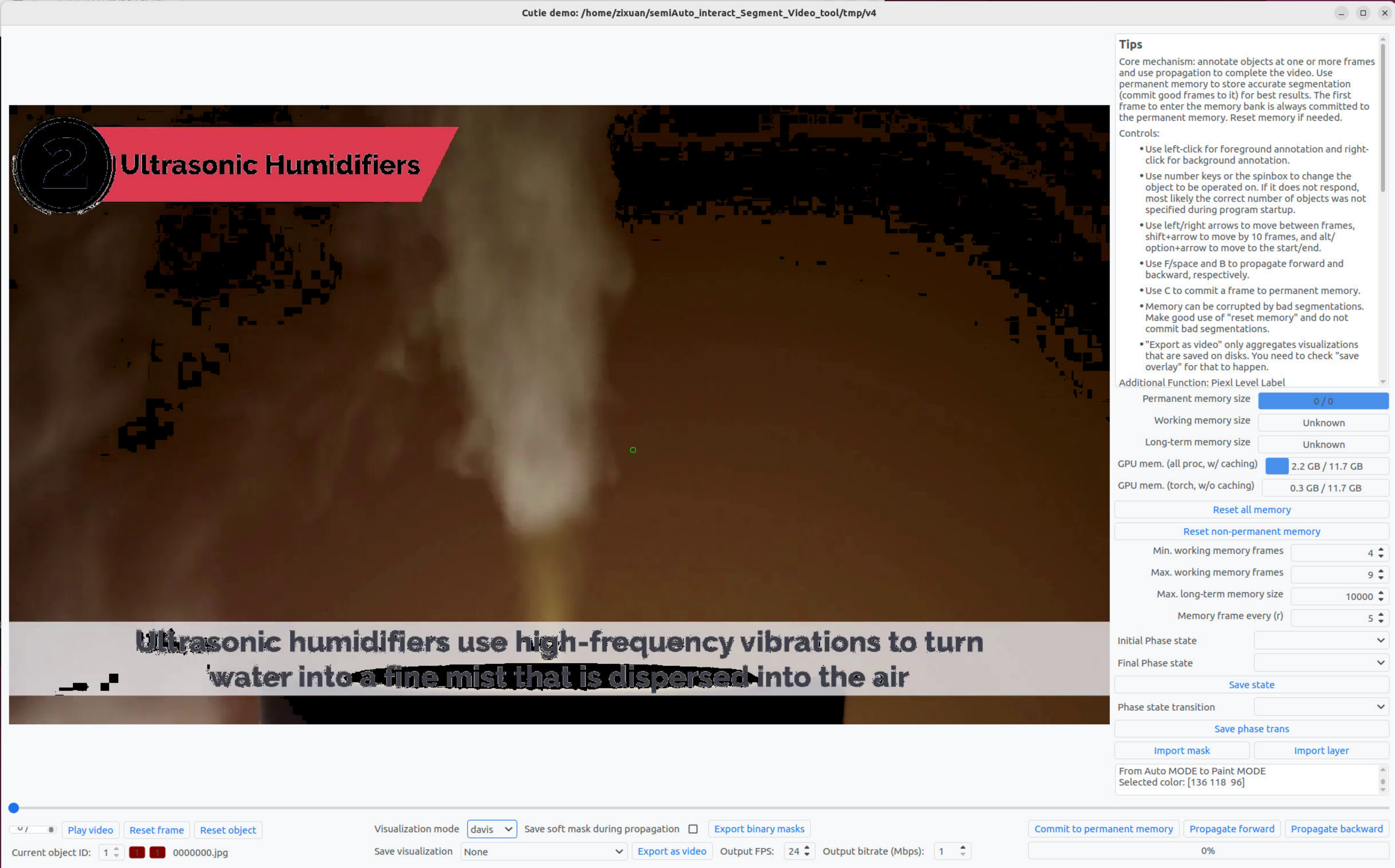}
        \caption{ \textbf{Appearance level}: select the pixel in the smoke with \textbf{high tolerance}. }
        \label{fig:apperance_level_a}
    \end{subfigure}
    \hfill
    \begin{subfigure}[b]{0.48\linewidth}
        \centering
        \includegraphics[width=\linewidth]{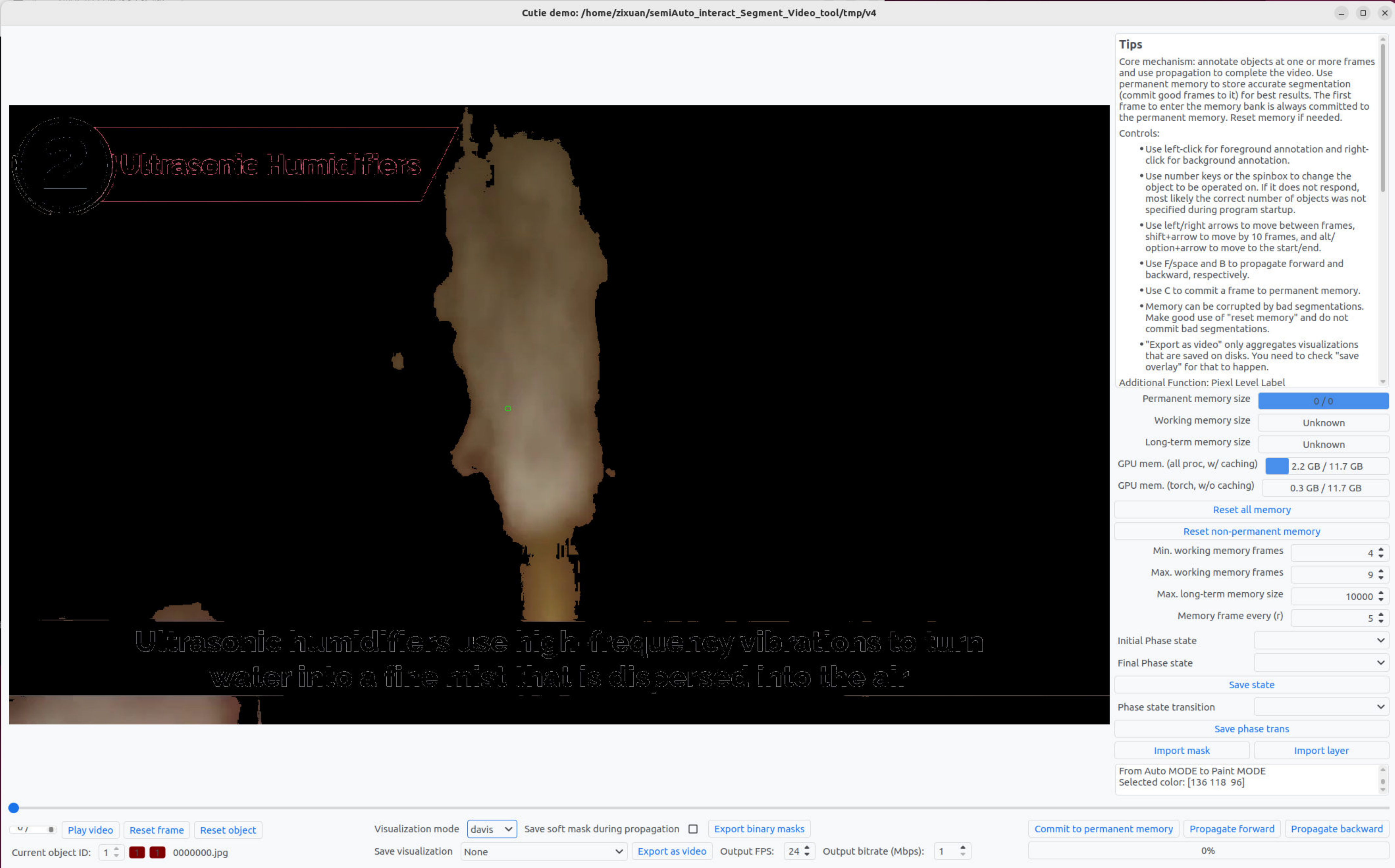}
        \caption{ \textbf{Appearance level}: select the pixel in the smoke with \textbf{low tolerance}. }
        \label{fig:apperance_level_b}
    \end{subfigure}

    \vspace{-5px}
    \caption{Multi-level semi-auto annotate tool: our annotate tool implements three-level annotating (\textbf{pixel level}, \textbf{appearance level}, \textbf{object level}). Using this tool, we can efficiently annotate different objects, such as small objects
\textit{garlic} in (a), 
solid with clear boundaries \textit{cola bottle} in (b) and fluid object \textit{smoke} in (c), (d). }
    \vspace{-5px}
    \label{fig:UI}
\end{figure*}

In \cref{fig:UI}, we show our details of the interactive UI of the multi-level semi-auto annotate tool. We implement this tool based on the interactive demo from Cutie~\cite{cheng2024putting}, including pixel level, appearance level, and object level. In particular, we implement the object-level function using the SAM2 model and Cutie model. In this way, we could perform the dual-model cross-validate analysis in \cref{sec:Model_Bias}.

\begin{table*}[t]
    \centering
    \caption{Details of core subset number of different Scenarios}
    \begin{tabular}{lcccc}
    \toprule
    \textbf{Scenario}    & \textbf{Full Set Number}   & \textbf{Core Subset Number}  & \textbf{Class Number}   & \textbf{Example}\\
    \midrule
    \textbf{Factory}     & 67                & 9                   & 13              & \textit{Disassemble/assemble a gun, Wrap a wire} \\
    \textbf{Handicraft}  & 40                & 12                  & 14              & \textit{Knit a sweater, Wrap a cigar}\\
    \textbf{Kitchen}     & 163               & 12                  & 70              & \textit{Cut celery, Shave fish}\\
    \textbf{Lab}         & 152               & 12                  & 56              & \textit{Drip liquid, Dissolve drug}  \\
    \textbf{Housework}   & 3                 & 3                   & 2               & \textit{Twist mop }\\
    \textbf{Decoration}  & 9                 & 2                   & 2               & \textit{Tear wallpaper} \\
    \textbf{Hospital}    & 7                 & 2                   & 4               & \textit{Ground herbal} \\
    \textbf{School}      & 1                 & 1                   & 1               & \textit{Sharpen a pencil} \\
    \textbf{Farm}        & 13                & 3                   & 7                & \textit{Shear a sheep} \\
    \textbf{Sport}       & 2                 & 1                   & 1                & \textit{Hit a balloon} \\
    \textbf{Daily live}  & 45                & 14                  & 18                & \textit{Pour tea, Shave beard}  \\
    \textbf{Experiment field} & 14           & 6                   & 3                  & \textit{Break glass, Twist a rubber}  \\
    \bottomrule
    \end{tabular}
    
    \label{tab:core_set}
\end{table*}

\section{Details of Core Subset} \label{sec:Core}
We extract a subset of cases that better represent the full dataset and refer to it as a core subset. For each specific scenario, we extracted a subset of cases. During the selection of the core subset in each scenario, we consider a series of factors: the number of the full set, the number of classes included, and the difficulty of cases. As is shown in \cref{tab:core_set}, We choose the size of the core subset of each scenario to make it closer to the proportion of the full set and the class number.

\section{Challenge Analysis} \label{sec:challenge}

In this part, we  explore how the size of the object and the velocity of the target object influence the performance of Cutie-ReVOS. 

\subsection{Definition of Object Size}

In our experiment, given a target object $o$ in the image $I$, its size is measured by the ratio between the mask of the object $M_{o}$ and the area of the image $A_{I}$, according to

\begin{equation}
    R(o) = \frac{M_{o}}{A_I},
\end{equation}
where $R(o)$ measures the relative size of the object compared to the Image. $M_o$ is the size of the ground-truth mask of the object $O$. $A_I$ is the area of Image $I$.

\subsection{Definition of Velocity}

Generally, the velocity of an object in an image is defined as the change in the centroid of the bounding box or mask per unit time. However, considering that we cannot measure the relationship between the distance in the image and the actual size of the object, we normalize the velocity based on the size of the object. Given a target object $o$ and the fps $f_v$ of the video clip, the relative velocity or the normalized velocity is defined as follows:

\begin{equation}
\begin{aligned}
        v(o) &= \frac{D(o) f_v }{M_o}, \\
    D(o) &= c_t(B_o) - c_{t-1}(B_o),
\end{aligned}
\end{equation}
where $B_o$ is the bounding box of target object $o$. $c_t(B)$ is the centroid of the bounding box in the timestamp $t$. $D(o)$ is the moving distance of the Object in a frame. 

\subsection{Relation between Challenge and Performance}

\begin{figure}[tbp]
    \vspace{-5px}
    \centering
    \begin{subfigure}[b]{0.47\linewidth}
        \centering
        \includegraphics[width=\linewidth]{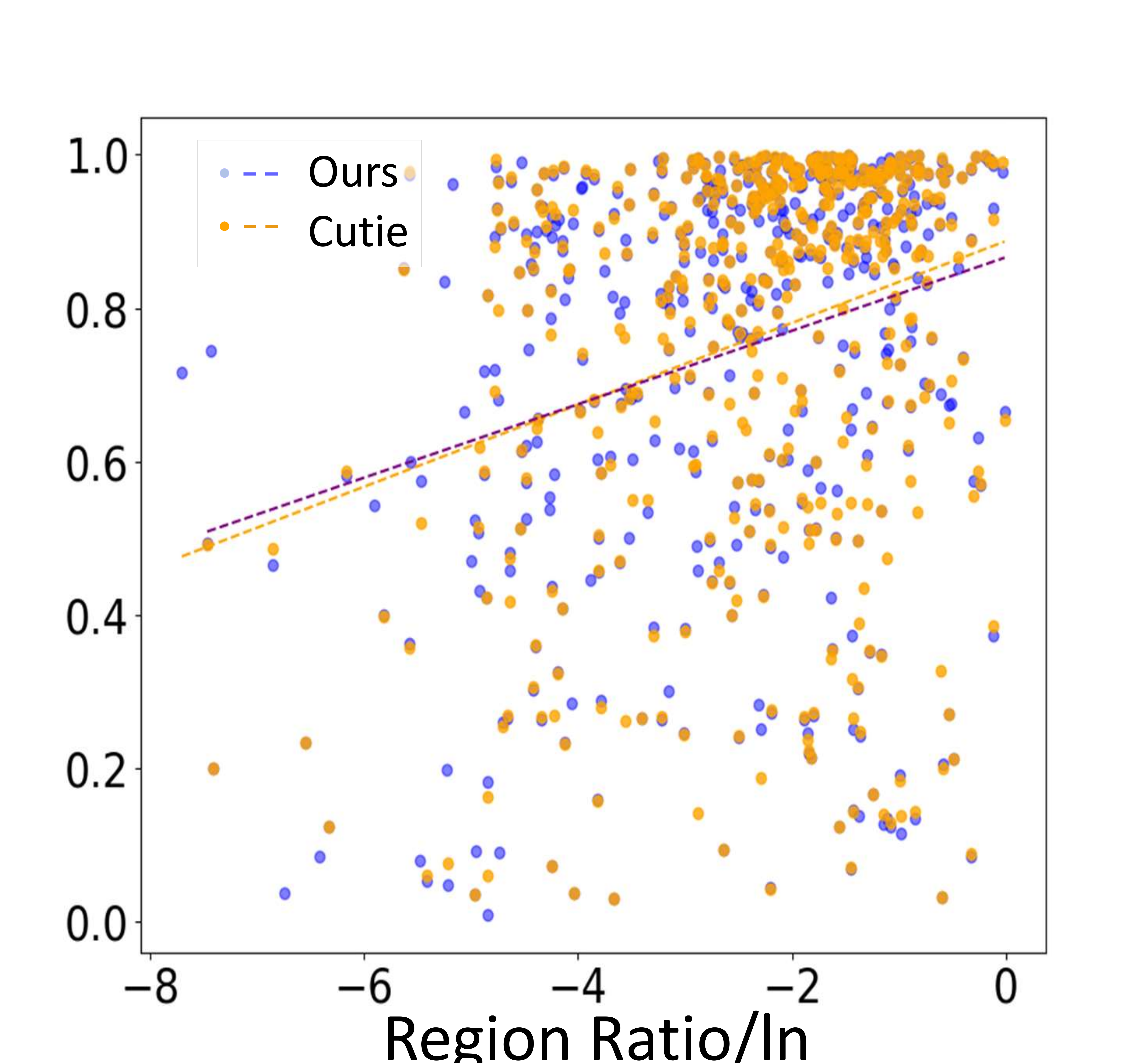}
        \caption{$\mathcal{J}_{mean}$-Region }
        \label{fig:J_and_size}
    \end{subfigure}
    \hfill
    \begin{subfigure}[b]{0.47\linewidth}
        \centering
        \includegraphics[width=\linewidth]{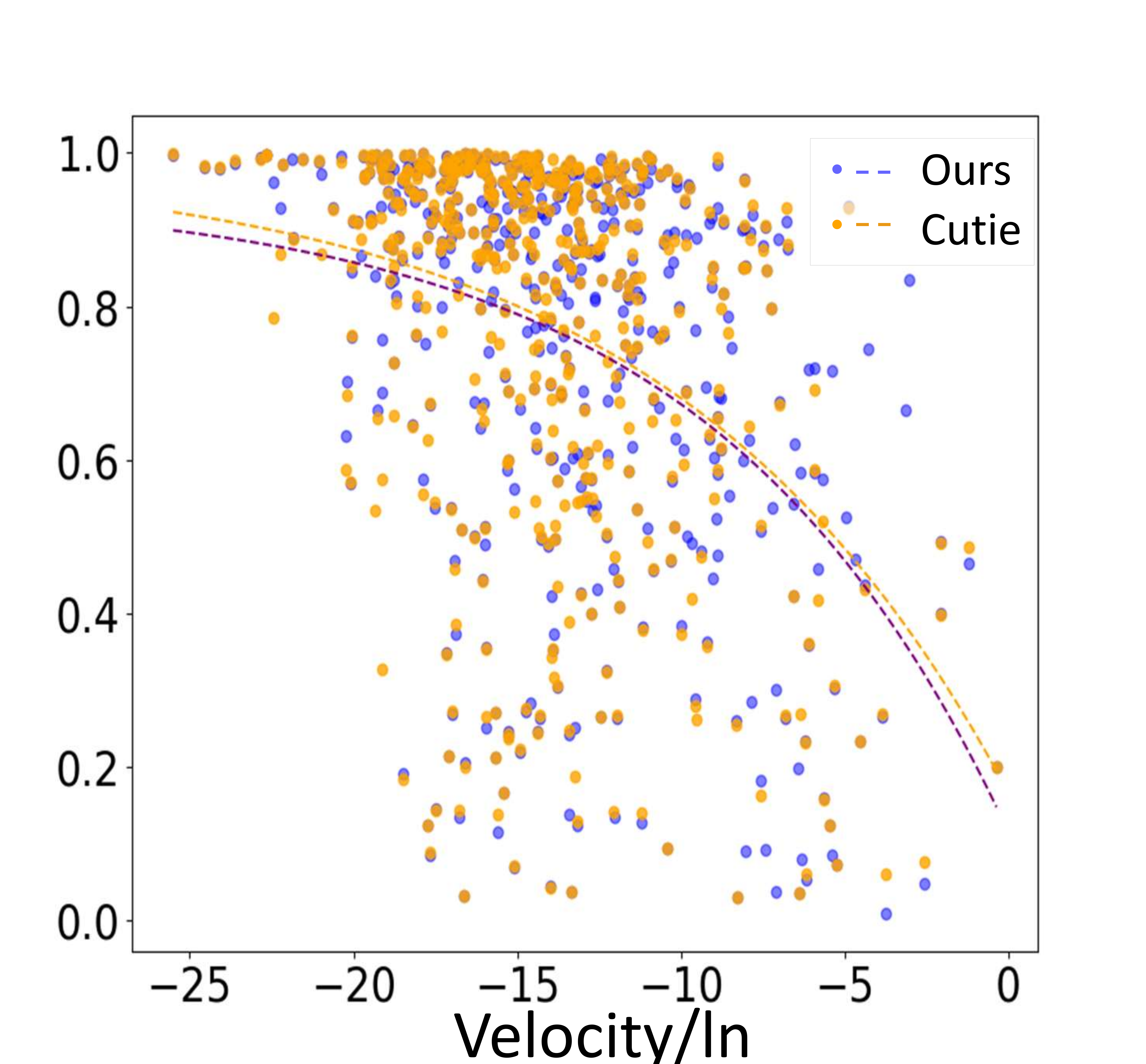}
        \caption{$\mathcal{J}_{mean}$-Velocity }
        \label{fig:J_and_v}
    \end{subfigure}
    \vspace{-5px}
    \caption{The relation between standard metric Jaccard index $\mathcal{J}_{mean}$ and region size, velocity.}
    \vspace{-5px}
    \label{fig:J_and_size_v}
 \end{figure}

As the curve in \cref{fig:J_and_size} demonstrated, the smaller the object's area ratio, the more challenging it is for the model to segment. Besides, for small objects, the performance of ReVOS-Cutie decreases compared to the original Cutie. For large objects, the situation is reversed. 

Similarly, the curve in \cref{fig:J_and_v}  indicates that  the relative velocity positively correlates with segmentation difficulty. We observe that when the velocity is more extreme, either too slow or too fast, the performance improvement of ReVOS-Cutie becomes more significant.

\section{More Failure Cases}\label{sec:failure}

\begin{figure*}[tbp]
    \vspace{-5px}
    \centering

        \centering
        \includegraphics[width=0.95\linewidth]{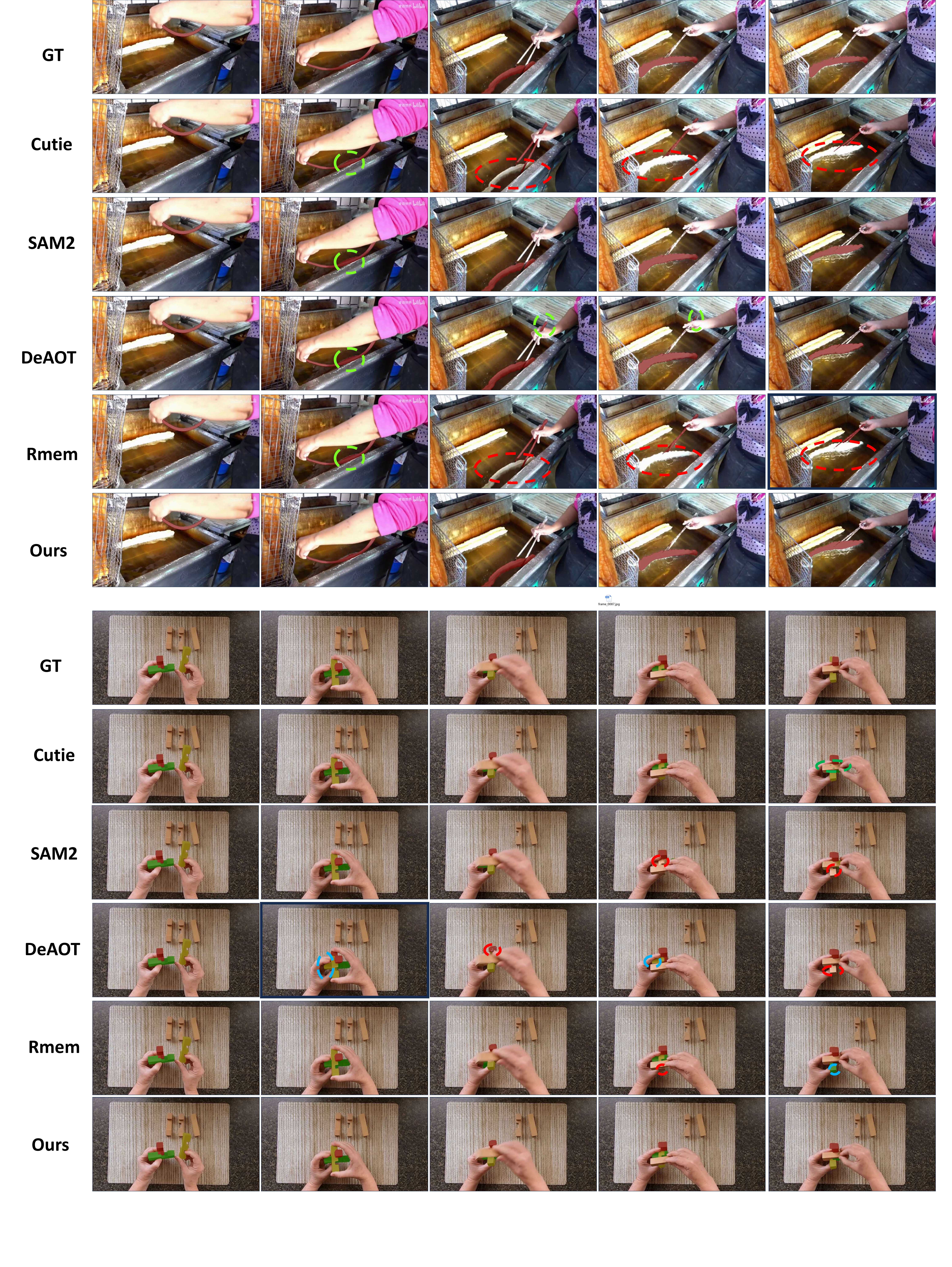}
    
    \vspace{-5px}
    \caption{Failure case 3 (\textit{fry dough}) and 4 (\textit{assemble puzzles}) in different models. (\textcolor{red}{Red circle}: false-negative region; 
   \textcolor{green}{Green circle}: false-positive region;
      \textcolor{blue}{Blue circle}: confuse-instance region). }

    \vspace{-5px}
    \label{fig:failure_case_2}
 \end{figure*}

\begin{figure*}[tbp]
    \vspace{-5px}
    \centering

        \centering
        \includegraphics[width=0.95\linewidth]{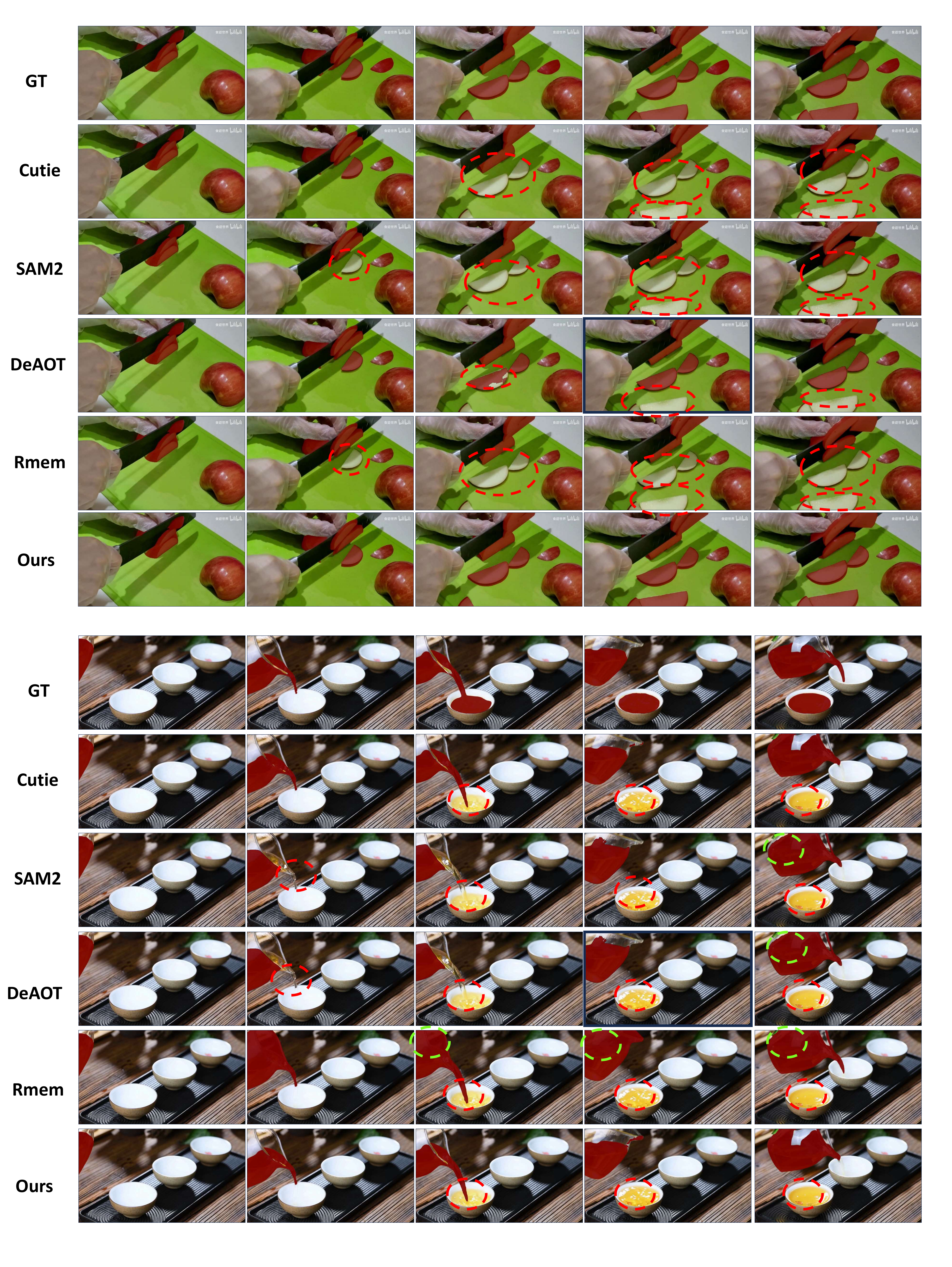}

    \vspace{-5px}
    \caption{Failure case 1 (\textit{cut apples}) and 2 (\textit{pour tea}) in different models. (\textcolor{red}{Red circle}: false-negative region; 
   \textcolor{green}{Green circle}: false-positive region). 
  } 
    \vspace{-5px}
    \label{fig:failure_case_1}
 \end{figure*}

In this part, we show more failure cases of the current models in \cref{fig:failure_case_1,fig:failure_case_2}. 

In case 1 (\textit{fry dough}) of \cref{fig:failure_case_2}, the boiling oil makes it difficult to separate the boundaries of the dough sticks accurately. Even for some models, the boiling oil causes the tracking loss. However, our method improves the segment accuracy  with visual distribution. 

In case 2 (\textit{assemble puzzles}) of  \cref{fig:failure_case_2}, the intra-solid transition with multi-instances usually instance confusion because of the similarity distribution. However, our methods are more robust when facing the intra-solid transition with multi-instances.

In case 1 (\textit{cut apples}) of \cref{fig:failure_case_1}, the tracking loss and mask incompleteness usually happen when the white pulp leaks out. The performance in the intra-solid phase transition with the color change challenge is not so good.

In case 2 (\textit{pour tea}) of \cref{fig:failure_case_1}, the flow of tea liquid into the tea cup is always accompanied by tracking loss. Besides, transparent teapots and tea liquids are confused and suffer from similar interference. Although our method improves this situation slightly, this multi-challenge case still has improved space.

\end{document}